\documentclass[lettersize,journal]{IEEEtran}
\usepackage{amsmath,amsfonts}
\usepackage{amsthm}
\usepackage{amssymb}
\usepackage[ruled,linesnumbered]{algorithm2e}
\usepackage{array}
\usepackage{bm}
\usepackage{booktabs}
\usepackage{overpic}
\usepackage{textcomp}
\usepackage{stfloats}
\usepackage{url}
\usepackage{verbatim}
\usepackage{graphicx}
\usepackage{cite}
\usepackage{multirow}
\usepackage{threeparttable}
\usepackage{flushend}

\usepackage{caption}
\usepackage{pifont}
\usepackage{subcaption}
\usepackage{color}
\usepackage{xcolor}
\usepackage{colortbl,booktabs}

\usepackage{tikz,siunitx}
\usetikzlibrary{shapes.geometric,shapes.symbols}

\usepackage[colorlinks,bookmarksopen,bookmarksnumbered,citecolor=green,linkcolor=red,urlcolor=green]{hyperref}

\newtheorem*{definition}{Problem Formulation}

\newcommand{\tablestyle}[2]{\setlength{\tabcolsep}{#1}\renewcommand{\arraystretch}{#2}\centering\footnotesize}
\newcommand{\cmark}{\ding{51}}%
\newcommand{\DAVIS}{$\textit{DAVIS}_{\textit{16}}$}
\newcommand{\FBMS}{\textit{FBMS}}
\newcommand{\STv}{\textit{SegTrackV2}}
\newcommand{\tikzsymbol}[3][circle]{\tikz[baseline=-0.5ex]\node[inner
	sep=#3,shape=#1,draw,#2]{};}%
	
\newcommand{\tablestar}{\tikzsymbol[star]{fill=black,star point ratio=2.618}{0.5pt}}
\newcommand{\tablecircle}{\tikzsymbol{fill=black}{0.8pt}}

\definecolor{tablered}{RGB}{205,51,51}
\definecolor{tablegreen}{HTML}{39b54a}
\definecolor{tableblue}{HTML}{4682B4}

\definecolor{figgrey}{RGB}{128,128,128}
\definecolor{figcyan}{RGB}{0,139,139}
\definecolor{figdarkred}{RGB}{64,0,0}
\definecolor{figred}{RGB}{205,51,51}
\definecolor{figgreen}{RGB}{113,198,113}

\definecolor{tomato}{HTML}{FF6347}
\definecolor{royalblue}{HTML}{4169E1}
\definecolor{springgreen}{HTML}{00FF7F}

\begin{document}

\title{Online Unsupervised Video Object Segmentation via Contrastive Motion Clustering}

\author{Lin~Xi,
	Weihai~Chen*,
	Xingming~Wu,
	Zhong~Liu,
	and Zhengguo~Li,
	\thanks{L. Xi, W. Chen, X. Wu, and Z. Liu are with the School of Automation Science and Electrical Engineering, Beihang University, Beijing 100191, China (e-mail: xilin1991@buaa.edu.cn; whchen@buaa.edu.cn; wxmbuaa@163.com; liuzhong@buaa.edu.cn).}
	\thanks{Z. Li is with the SRO Department, Institute for Infocomm Research, Agency for Science, Technology and Research (A*STAR), 1 Fusionopolis Way, \#21-01, Connexis South Tower, Singapore 138632, Republic of Singapore (e-mail: ezgli@i2r.a-star.edu.sg).}
	\thanks{*Corresponding author: Weihai Chen.}}



\maketitle

\begin{abstract}
	Online unsupervised video object segmentation (UVOS) uses the previous frames as its input to automatically separate the primary object(s) from a streaming video without using any further manual annotation. A major challenge is that the model has no access to the future and must rely solely on the history, \emph{i.e.}, the segmentation mask is predicted from the current frame as soon as it is captured. In this work, a novel contrastive motion clustering algorithm with an optical flow as its input is proposed for the online UVOS by exploiting the common fate principle that visual elements tend to be perceived as a group if they possess the same motion pattern. We build a simple and effective auto-encoder to iteratively summarize non-learnable prototypical bases for the motion pattern, while the bases in turn help learn the representation of the embedding network. Further, a contrastive learning strategy based on a boundary prior is developed to improve foreground and background feature discrimination in the representation learning stage. The proposed algorithm can be optimized on arbitrarily-scale data (\emph{i.e.}, frame, clip, dataset) and performed in an online fashion. Experiments on \DAVIS, \FBMS, and \STv\ datasets show that the accuracy of our method surpasses the previous state-of-the-art (SoTA) online UVOS method by a margin of 0.8\%, 2.9\%, and 1.1\%, respectively. Furthermore, by using an online deep subspace clustering to tackle the motion grouping, our method is able to achieve higher accuracy at $3\times$ faster inference time compared to SoTA online UVOS method, and making a good trade-off between effectiveness and efficiency. Our code is available at \href{https://github.com/xilin1991/ClusterNet}{https://github.com/xilin1991/ClusterNet}.
\end{abstract}


\section{Introduction}
\label{sec:intro}

\begin{figure}[t]
	\begin{center}
		\includegraphics[width=1.0\linewidth]{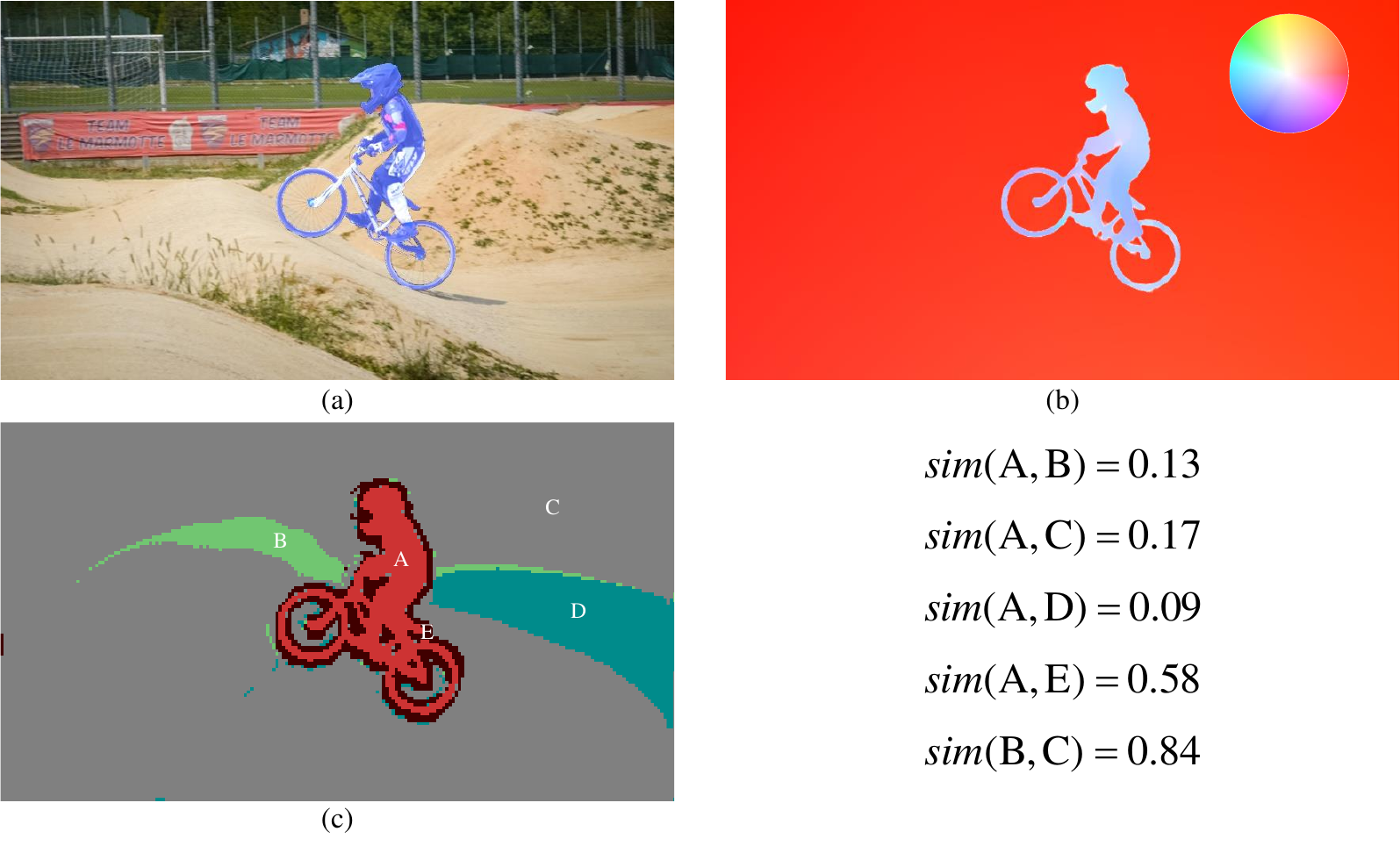}
	\end{center}
	\vspace{-3pt}
	\caption{Motion grouping. (a) The RGB image with ground truth; (b) the optical flow visualized by the inset color wheel; (c) the motion segmentation using our proposed prototypical subspace clustering framework (clusters $k$=5). According to the prototypical subspace bases, we clustered different motion patterns (\emph{i.e.}, A-\textcolor{figred}{\textbf{red}}, B-\textcolor{figgreen}{\textbf{green}}, C-\textcolor{figgrey}{\textbf{grey}}, D-\textcolor{figcyan}{\textbf{cyan}}, and E-\textcolor{figdarkred}{\textbf{dark red}}). $\textit{sim}(\cdot)$ is the similarity of different motion patterns and is normalized to $[0, 1]$.}
	\label{fig:Head}
\end{figure}

\IEEEPARstart{W}{hen} looking around in a dynamic scene, visual elements moving at the same speed and/or direction tend to attract human attention as part of a single stimulus. This principle is called common fate and is theorized by Gestalt psychology \cite{Gestalt}. A common example is a BMX rider going through dirt jumps. If the rider and the BMX bike have the same trajectory, they are perceived as the same motion group. The background, which has a different trajectory than the BMX rider, does not appear to be part of the same group, as shown in Fig. \ref{fig:Head}.

According to the Gestalt principle of common fate, objects should share a common destination, moving together consistently throughout the scene. Therefore, the motion of objects can serve as an important cue for video object segmentation (VOS). In computer vision, the pixel-wise motion in the scene can be obtained from an optical flow estimation and used to determine, segment, and learn the objects. In the recent literature of VOS, many learning-based models \cite{LVO, LSMO, FSEG, MotAdapt, MATNet, EpONet, AGS, AGS_TPAMI, COSNet, COSNet_TPAMI, AGNN, AGNN_TPAMI, FEMNet, IMCNet, RTNet, TransportNet} have been proposed to learn more discriminative objectness by leveraging motion information. While such models through supervised learning require massive pixel-wise annotations, they are limited to a small range of object categories predefined in the datasets. To reduce the cost of data labeling, numerous unsupervised approaches \cite{ELM, FST, NLC, SAGE, STP, CIS, Traj} have been proposed that use the motion cues. However, traditional physics-optimization-based approaches incur significant computational costs due to the optimization process over the entire video. Unsupervised methods \cite{DyStaB, MG, Sprites} based on deep neural networks have gained significant advantage by learning a deep representation on a video dataset. Although those methods achieve good performance, they cannot handle streaming video because they work offline and require the entire video to be processed before making predictions.

Unlike unsupervised video object segmentation (UVOS) in the offline setting, a major challenge for online UVOS is that the inference is performed solely on observations of the past, without utilizing the information from video frames in the future. While processing video online is beneficial for many applications, such as video compression \cite{VidCom1, VidCom2, VidCom3, VidCom4}, analysis \cite{VidAnalsis, HumanParsing1, HumanParsing2, HumanParsing3, HumanParsing4}, and editing \cite{VidEdt}. Tokmakov \emph{et al.} \cite{LMP} proposed an end-to-end network to map the optical flow to motion segmentation, followed by an object proposal model \cite{PinheiroLCD16} to extract the candidate objects. Similarly, Zhou \emph{et al.} \cite{UOVOS} proposed a method based on salient motion detection and object proposals which were directly predicted by a pre-trained model \cite{MaskRCNN} without fine-tuning to obtain final CRF refined results. While these methods require training on a large dataset or object masks, they incur computational cost in the optimization (training) and inference process. In addition, some online UVOS methods suffered from another shortcoming: inappropriate use of motion cues. For example, Taylor \emph{et al.} \cite{CVOS} used long-term temporal information to initialize the target object on a few frames for online unsupervised framework. However, uninformative history frames cause error accumulation during the propagation, which degrades the performance of online UVOS. Perazzi \emph{et al.} \cite{SFM} employed a salient detection method without considering the object information and motion cues of video content, which is a limitation in UVOS where the segmented object was defined as the main object in the scene with a distinctive motion.

We argue that an online UVOS model must predict frame-level segmentation in correlation to what happened hitherto, and as efficiently as possible for the online setting. Therefore, an online UVOS model should satisfy the following criteria: (i) frame-by-frame manner, (ii) relatively fast optimization and inference, and (iii) short-term temporal dependence.

\begin{figure}[t]
	\begin{center}
		\includegraphics[width=1.0\linewidth]{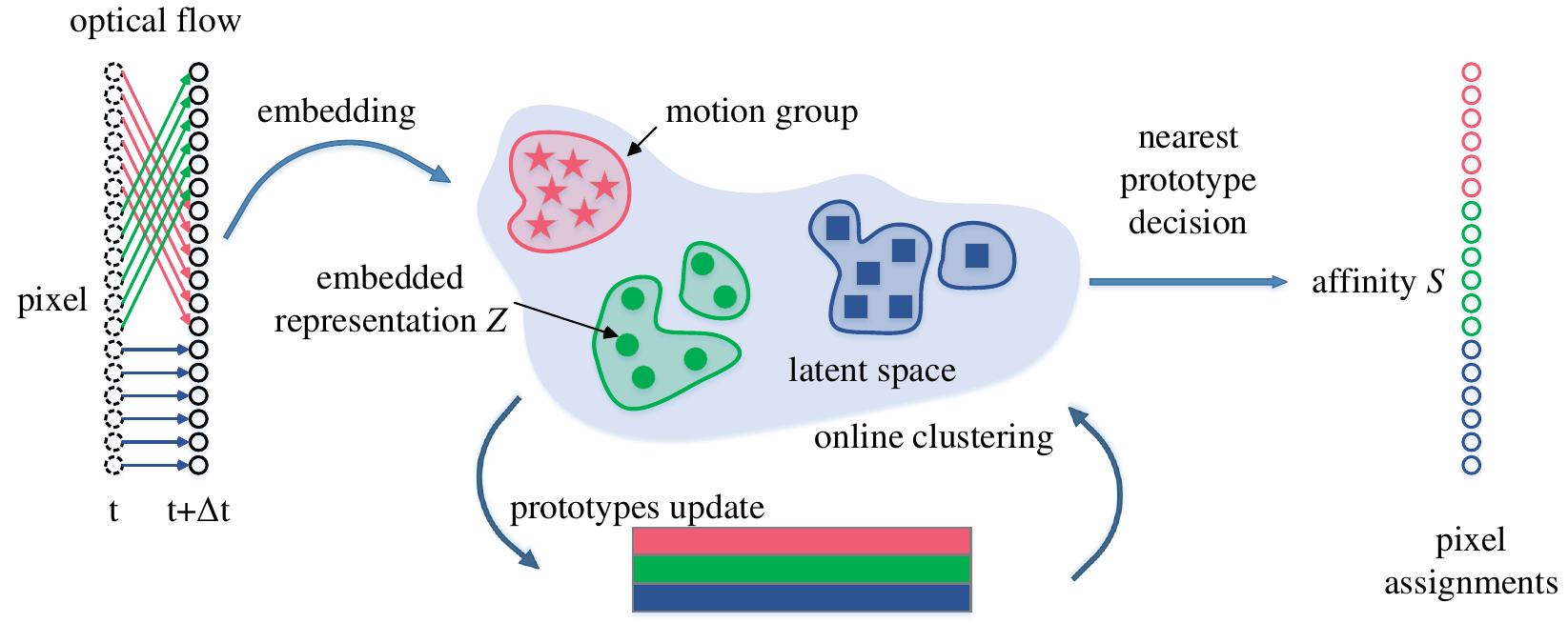}
	\end{center}
	\vspace{-3pt}
	\caption{Paradigm of the proposed clustering method. We iteratively summarize prototypical bases from the embedded representation $\bm{Z}$, and the $\bm{Z}$ are then combined with the prototypes to compute the affinity $S$.}
	\label{fig:draft}
\end{figure}

In this paper, a novel online UVOS algorithm is proposed by using an efficient and accurate online deep subspace clustering for motion grouping, which directly factorizes the optical flow into $k$ groups corresponding to $k$ subspaces. The proposed algorithm processes one video frame at a time without any additional pre/post-processing (\emph{i.e.}, \cite{Yilun, CRF}), and the results for the frame is depend only on the previous frame. A simple video auto-encoder model is designed to summarize a set of subspace prototypes from the latent space, where the deep auto-encoder is optimized on each individual video sequence and does not require training on a large dataset. Specifically, the motion segmentation problem is addressed by training a generative model that is used to learn an embedded representation $\bm{Z}$ of the optical flow. The $\bm{Z}$ is then combined with the centers of each segment, \emph{i.e.}, the prototypes, to construct the subspace affinity vector $\bm{S}$ for pixel assignments. Each pixel is assigned to the nearest prototype without relying on additional learnable parameters. The prototypes are formed by clustering nearby points in the embedding space. They represent motion groups following the common fate principle. The network structure of the proposed method is illustrated in Fig. \ref{fig:draft}. In order for the auto-encoder to effectively learn the discriminative features between the foreground and background, we further exploit an important scene prior \cite{SP}, \emph{i.e.}, the optical flow at the boundary of an image is significantly different from the motion direction of the object of interest, to design a pixel-level contrastive learning strategy.

Overall, our main contributions are: 1) a novel online deep subspace clustering method for the online UVOS by exploiting the motion cue; 2) an effective optimization approach for the online clustering that can handle an arbitrary video independently without being trained on large datasets; and 3) a pixel-level contrastive learning strategy that significantly improves the foreground and background feature discrimination for the auto-encoder. To validate these contributions, the proposed method is evaluated on three public benchmarks (\emph{i.e.}, \DAVIS, \FBMS, and \STv); the proposed algorithm outperforms state-of-the-art (SoTA) online UVOS models, while being faster to optimize and infer.

\section{Related works}
\label{sec:related}

\noindent \textbf{Motion segmentation} is to identify and segment independently moving objects in a video, that is, to solve the problem of motion grouping. Many approaches tackle the issue from a motion clustering point of view. Shi \emph{et al}. considered motion segmentation as a spatio-temporal image clustering problem \cite{NorCuts}. To increase robustness, some methods use motion cues, such as point trajectories \cite{Point1, Point2, Point3, Point4} or optical flow \cite{OP1, FST} accumulated over multiple frames, to segment moving objects. Luo \emph{et al}. \cite{Traj} proposed a complexity awareness framework that exploits local clips and their relationships for motion segmentation. Kumar \emph{et al}. proposed an algorithm to obtain the initial estimate of the model by dividing the scene into rigidly moving components to solve a grouping problem to associate pixels into a number of motion clusters \cite{Kumar}. Brox \emph{et al}. defined pairwise distances between point trajectories from adjacent frames for the motion clustering \cite{Brox}. Ochs and Brox \cite{Ochs} adopted the spectral clustering on hypergraphs which is a similarity map obtained from a third motion vector, instead of pairs \cite{Brox} to segment point trajectories.

It is worth noting that Xie \emph{et al}. \cite{Xie} also inserted motion clustering into their object segmentation problem pipeline. However, under the premise of a supervised setting, this method introduces a pixel-trajectory recurrent neural network that learns the trajectories of foreground pixels and clusters pixels over time. In contrast, the proposed algorithm learns motion patterns only using the optical flow without requiring any manual annotation. In addition, our resulting feature representations which are optimized on individual videos give us a global dependence over entire videos.

\begin{figure*}[htbp]
	\begin{center}
		\includegraphics[width=0.9\textwidth]{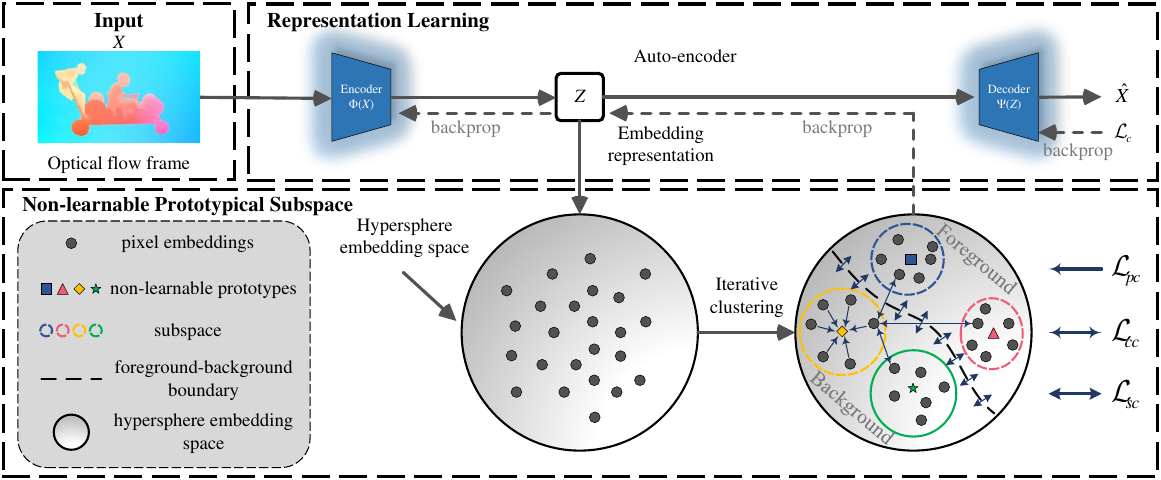}
	\end{center}
	\vspace{-3pt}
	\caption{The overview optimization diagram for our proposed method with the optical flow as our input. Given an optical flow $\bm{X}$, we utilizes auto-encoder to embed it into a $p$-dimensional embedding feature $\bm{Z}$ and outputs its corresponding reconstruction $\bm{\hat{X}}$. During the optimization phase, we iteratively summarize non-learnable prototypical bases for the motion pattern, while the bases are constrained by our proposed contrastive learning strategy to help shape the feature space. To obtain the final cluster labels, we use the proposed subspace clustering algorithm with a hard assignment to group each pixel to the prototypical bases.}
	\label{fig:framework}
\end{figure*}

\noindent\textbf{Unsupervised video object segmentation} aims to automatically identify and segment the most visually prominent objects from the background in sequences, unlike semi-supervised \cite{OSVOS, MaskTrack, STM, STCN, XMem} and referring \cite{MLRLSA_RVOS, VLT, URVOS} video object segmentation which involves human inspection. Recently, many approaches \cite{LVO, LSMO, SFL, LMP, PDB} have been proposed to tackle the offline UVOS. Although the term ``unsupervised'' is used here, in practice there are some differences from \textit{fully unsupervised} settings. In general, many popular algorithms \cite{FSEG, MATNet, EpONet, AGS, AGS_TPAMI, COSNet, COSNet_TPAMI, AGNN, AGNN_TPAMI, FEMNet, IMCNet, TransportNet, RTNet} require supervised training on large-scale datasets to obtain the segmentation masks. Alternatively, a number of works \cite{MG, DyStaB, Sprites} based on the offline setting employ a deep neural network to discover the objects of interest from the perspective of completely unsupervised concepts in the traditional methods. Lu \emph{et al.} \cite{MuG} proposed a unified framework for unsupervised learning, aimed at object segmentation through the exploitation of the inherent consistency across adjacent frames in unlabeled videos. Similar to our work, Yang \emph{et al}. \cite{MG} used slot attention \cite{Slot} to learn to segment objects in a self-supervised manner, and also took the optical flow as the input of the auto-encoder, which is a type of generative model (\emph{e.g.}, VAEs \cite{VAE} and GANs \cite{GAN, DSRGAN}), but slot attention and binding graph neural networks (GNN) rely on large-scale datasets for training. DyStaB employs static and dynamic models to learn object saliency from motion in a video, which can then be applied at inference time to segment objects, even in static images \cite{DyStaB}. Deformable Sprites (DeSprites) \cite{Sprites} are a type of video auto-encoder model that is optimized on each individual video. Our work also optimizes an auto-encoder on a specific sequence in an unsupervised manner. Unlike the SoTA offline UVOS method DeSprites, our goal is to cluster the points that share motion patterns in the embedding space, which significantly improves the effectiveness of the optimization and reduces the inference time for online UVOS.

\noindent\textbf{Subspace clustering} is to segment the original data space into its corresponding subspace. Classical subspace clustering methods used kernels to transform the original data into a high-dimensional latent feature space in which subspace clustering is performed \cite{CSCN}. Recently, there have been a few works that used deep learning techniques for feature extraction in subspace clustering. Ji \emph{et al}. developed a convolutional auto-encoder network combined with a self-expression module \cite{DSCN}, which showed significant improvement on several image datasets.  Instead of constructing the affinity matrix for subspace clustering, Zhang \emph{et al}. utilized deep neural networks to iteratively project data into a latent space and update the k-subspaces \cite{kSCN}. In \cite{kFSC1, kFSC2}, a k-factorization subspace clustering was proposed for large-scale subspace clustering, which effectively reduces the complexity of clustering.

In this paper, we attempt to develop a joint optimization framework for the online UVOS that can simultaneously learn feature representation and subspace clustering. Inspired by the effectiveness of the k-FSC model \cite{kFSC1}, we combine a powerful CNN to define $k$ non-learnable prototypes in the latent space as the $k$-subspace of clustering.

\noindent\textbf{Contrastive learning} is an attention-grabbing unsupervised representation learning method that maximizes the similarity of positive pairs while minimizing the similarity of negative pairs in a feature space \cite{Contrastive1, Contrastive2}.  Li \emph{et al}. proposed the contrastive clustering method, which performs dual contrastive learning at the instance and cluster level under a unified framework \cite{CC}. By adopting the foreground-background saliency prior \cite{SP} for contrastive learning, we propose a novel pixel-level contrastive learning framework without the requirement of image-level supervision. Features from the foreground are pulled together and contrasted against those from the background, and vice versa.

\section{Methods}
\label{sec:methods}

In our ``online unsupervised'' setting, an optical flow is taken as our input and all pixels are assigned into different groups to predict a segment containing the moving object by an online deep subspace clustering on top of non-learnable prototypes. One video frame at a time is processed, and the results for the frame depend only on the previous frame. The overall framework of the proposed method is shown in Fig. \ref{fig:framework}.

\begin{definition}\label{def:problem}
	Let $\{\bm{X}_{t\rightarrow t+\Delta t}^{(i)} \in \mathbb{R}^{H \times W \times 2}\}_{i=1}^{N}$ ($N\in\mathbb{N}^{*}$) denote optical flow frames from the individual video, where $H \times W$ indicates the spatial resolution of images. Assume that a pixel point $\mathbf{x}_{t\rightarrow t+\Delta t}^{(s)}$ ($s \in \mathbb{R}^{H \times W}$) in an optical flow frame is drawn from a $p$-dimensional subspaces $\{\mathcal{S}_{j}\}_{j=1,\cdots,k}$ (i.e., $\mathbf{x}_{t\rightarrow t+\Delta t}^{(s)} \in \mathcal{S}_{j}$), where $k$ clusters correspond to $k$ different subspaces. For $j=1,...,k$, the pixel point $\mathbf{x}_{t\rightarrow t+\Delta t}^{(s)}$ can be formally represented as:
	\begin{equation}\label{eq:1}
		\mathbf{x}_{t\rightarrow t+\Delta t}^{(s)}=g(\bm{U}_{j}\mathbf{v})+\epsilon,
	\end{equation}
	where one subspace can be expressed by a specific subspace base $\bm{U}_{j}\in\mathbb{R}^{p\times r_{j}}$ ($p<r_{j}$), $\mathbf{v}\in\mathbb{R}^{r_{j}}$ denotes a random variable, $\epsilon \in \mathbb{R}^{2}$ is random noise, and $g:\mathbb{R}^{p}\rightarrow\mathbb{R}^{2}$. Find $k$ cluster patterns from the $r$-dimensional latent space $\mathcal{U}$ and $p<r$.
\end{definition}

\subsection{Network Formulation}

The auto-encoder is a widely used self-supervised model and it can embed the raw data into a customizable latent space. A network architecture consisting of multiple convolutional layers is adopted to map the optical flow into the $r$-dimensional latent space $\mathcal{U}$, and then the $p$-dimensional embedding features $\bm{Z}\in\mathbb{R}^{\frac{H}{c}\times\frac{W}{c}\times p}$ is denoted by
\begin{equation}\label{eq:2}
	\bm{Z}=F(\Phi(\bm{X}))+A(F(\Phi(\bm{X}))),
\end{equation}
where $c$ is a scale, and a feature multi-layer perceptron (MLP) which is stacked after the embedding features $\Phi(\bm{X})$ is denoted by $F:\mathbb{R}^{r}\rightarrow\mathbb{R}^{p}$ ($p<r$). A spatial attention module $A(\cdot)$, which is implemented by the sum of max and average pooling followed by an upsampling layer, is added to improve the spatial stability. 

The embedding features $\bm{Z}$ are fed into a decoder $\Psi(\cdot)$ to reconstruct the optical flow, and the reconstruction loss $\mathcal{L}_{c}$ for the auto-encoder is formulated as:
\begin{equation}\label{eq:3}
	\mathcal{L}_{c}=\frac{1}{S}\parallel\bm{X}-\bm{\hat{X}}\parallel_{F}^{2}=\frac{1}{S}\sum_{s\in S}\parallel\mathbf{x}^{(s)}-\mathbf{\hat{x}}^{(s)}\parallel_{2}^{2}
\end{equation}
where $\bm{\hat{X}}=\Psi(\bm{Z})$ and $S$ is the entire spatial grid. For simplicity, the subscript $t\rightarrow t+\Delta t$ is omitted for the optical flow ward in Eqs. \ref{eq:2} and \ref{eq:3}. We only leverage the temporal information from the previous frames for the optical flow estimation, hence $\Delta t < 0$ in the online setting.

\subsection{Non-learnable Prototypical Subspace Clustering}

\begin{figure}[!htbp]
	\begin{center}
		\includegraphics[width=0.6\linewidth]{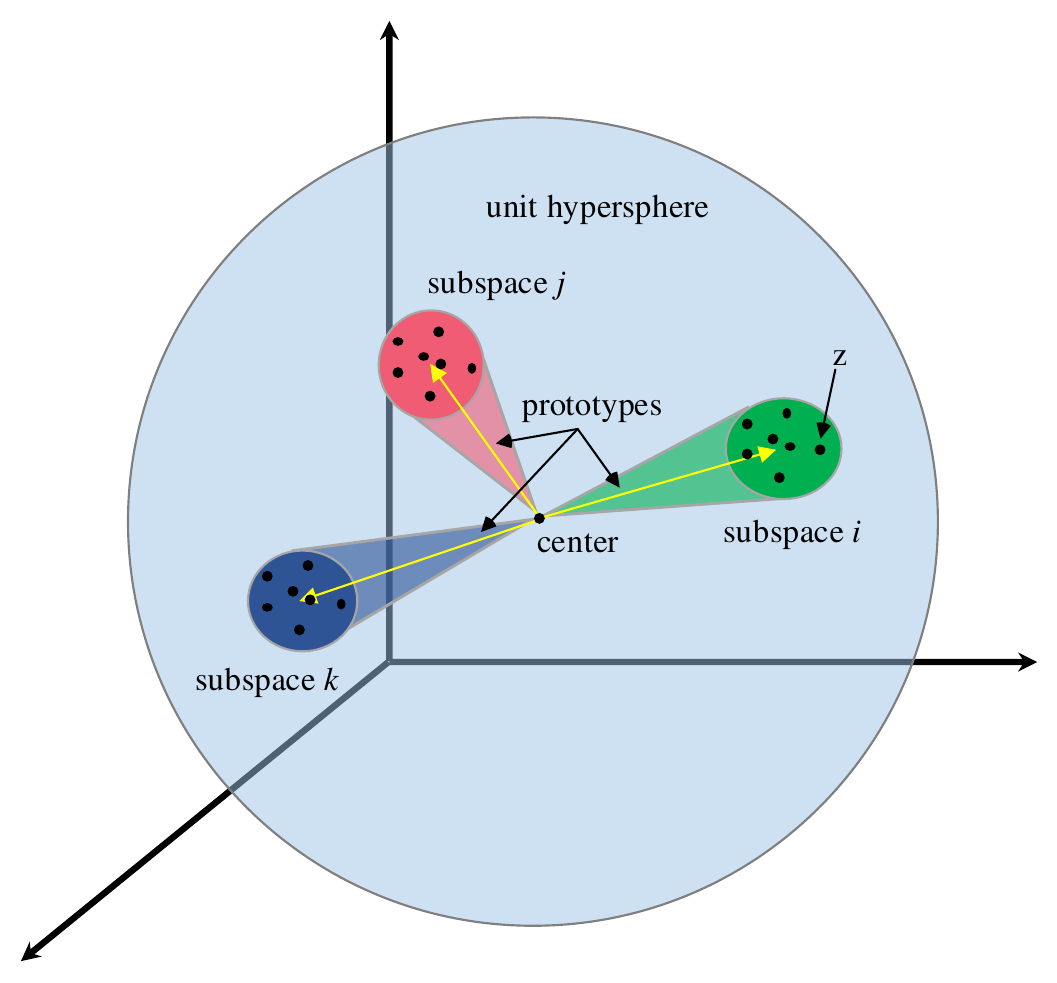}
	\end{center}
	\vspace{-3pt}
	\caption{The simplified diagram of $p-1$ dimensional unit hypersphere, where each subspace corresponds to the surface area of the unit hypersphere centered on different prototypes, denoted as $\mathcal{P}_{j}$. When $\parallel\mathcal{P}_{i}^{\top}\mathcal{P}_{j}\parallel$ is sufficiently small for all $i\neq j$, it means that each prototype $\mathcal{P}_{j}$ on the unit hypersphere is situated at a greater distance, enabling the identification of a suitable boundary for clustering.}
	\label{fig:hypersphere}
\end{figure}

Suppose $\bm{P}^{\top}\bm{X}=\bar{\bm{X}}$, where $\bar{\bm{X}}=[\bar{\bm{X}}_{1},\bar{\bm{X}}_{2},\cdots,\bar{\bm{X}}_{k}]$ and $\bm{P}$ is an unknown permutation matrix. According to the assumption in the proposed \textbf{Problem Formulation}, once $\bm{U}_{j}$ is obtained from Eq. \ref{eq:1}, the correct clusters are then identified. This implies that the $\bm{U}_{j}$ is not explicitly determined. Instead, a neural network is exploited to replace Eq. \ref{eq:1} by approximating $\mathbf{x}^{(s)}$, which yields the following formulation:
\begin{equation}\label{eq:4}
	\hat{\mathbf{x}}^{(s)}=\Psi(\hat{\bm{U}}_{j}\hat{\mathbf{v}^{(s)}}),
\end{equation}
where $\mathbf{z}^{(s)}:=\hat{\bm{U}}_{j}\hat{\mathbf{v}^{(s)}}$ ($\mathbf{z}^{(s)}\in\mathbb{L}^{(s)}$), $\mathbf{z}^{(s)}$ refers to the embedding feature associated with pixel $\mathbf{x}^{(s)}$, and $\mathbb{L}^{(s)}$ indicates the true cluster to which $\mathbf{x}^{(s)}$ should be assigned. There exists a direct correspondence between the spatial positions of embedding features and pixels, establishing a one-to-one relationship between $\mathbf{z}^{(s)}$ and $\mathbf{x}^{(s)}$. In fact, it is difficult to determine $\mathbb{L}^{(s)}$ directly. Now the embedding feature $\mathbf{z}^{(s)}$ is $\ell_{2}$ normalized so that it lies on the surface of a unit hypersphere, as shown in Fig. \ref{fig:hypersphere}. A new variable $\mathcal{P} \in \mathbb{R}^{p\times k}$ is introduced as the subspace prototypes, where $\mathcal{P}=[\mathcal{P}_{1},\mathcal{P}_{2},\cdots,\mathcal{P}_{k}]$ and $\parallel\mathcal{P}_{j}\parallel$=1, $j=1,\cdots,k$, and $p$ is the dimension of each prototype. The function of $\mathcal{P}_{j}$ is to summarize the subspace $\mathcal{S}_{j}$, $j=1,\cdots,k$. Thus $\parallel\mathcal{P}_{i}^{\top}\mathcal{P}_{j}\parallel$ is assumed to be small enough for all $i\neq j$, \emph{i.e.},
\begin{equation}\label{eq:5}
	\parallel\mathcal{P}_{i}^{\top}\mathcal{P}_{j}\parallel\leq\tau,\ i\neq j,
\end{equation}
where $\tau$ is a small constant.

The embedding feature $\mathbf{z}^{(s)}$ of a data sample is compared with $\mathcal{P}_{j}$ ($j=1,\cdots,k$) to obtain the winning prototype as
\begin{equation}\label{eq:6}
	\alpha^{(s)}=\arg\max_{j}\parallel {\mathbf{z}^{(s)}}^{\top}\mathcal{P}_{j}\parallel.
\end{equation}
It is assumed that
\begin{equation}\label{eq:7}
	s^{(s,\alpha)}=\parallel {\mathbf{z}^{(s)}}^{\top}\mathcal{P}_{\alpha^{(s)}}\parallel\gg\max_{j\neq\alpha^{(s)}}\parallel{\mathbf{z}^{(s)}}^{\top}\mathcal{P}_{j}\parallel,\ s=1,\cdots,S,
\end{equation}
where $s^{(s,\alpha)}$ denotes the affinity. In other words, maximizing the likelihood of Eq. \ref{eq:6} is assigning $\mathbf{z}^{(s)}$ to one of $\mathcal{P}$ with a probability distribution:

\begin{equation}\label{eq:8}
	p(\alpha^{(s)}|\mathbf{z}^{(s)})=\frac{\exp(s^{(s,\alpha)})}{\sum_{j=1}^{k}\exp(s^{(s,j)})}.
\end{equation}

An online clustering strategy is adopted to update $\alpha^{(s)}$ so that the pixels with the same motion pattern are assigned to the prototype $\mathcal{P}_{j}$ belonging to that subspace $\mathcal{S}_{j}$ according to $s^{(s, j)}$. It can be known from the permutation matrix $\bm{P}$ that the mapping $\bm{T}$ that assigns the pixel $\mathbf{x}^{(s)}$ to the prototypes is related to it as
\begin{equation}\label{eq:9}
	\bm{T}^{\top}\mathcal{P}^{\top}\bm{Z}\propto \bm{P}^{\top}\bm{X},
\end{equation}
where the column of $\bm{T}\in\mathbb{R}^{k\times p}$ is the one-hot assignment vector of pixel $\mathbf{x}^{(s)}$ over $k$ prototypes. In other words, each pixel is assigned to a single prototype, and the sum of pixels matched by all the prototypes is equal to all pixels in the frame. Thus, the augmented assignment $\bm{T}$ now has the following constraints:
\begin{equation}\label{eq:10}
	\bm{T}^{\top}\bm{1}_{k}=\bm{1}_{S}\ \textrm{and}\ \bm{T}\bm{1}_{S}=\frac{S}{k}\bm{1}_{k},
\end{equation}
where $\bm{1}_{k}$ denotes the vector of all ones of $k$ dimensions.

The mapping $\bm{T}$ can be optimized by maximizing the probability distribution $p(\alpha^{(s)}|\mathbf{z}^{(s)})$ (Eq. \ref{eq:8}) between the pixel embedding $\bm{Z}$ and the prototypes $\mathcal{P}$. The solution of the above optimization problem corresponds to the optimal transport \cite{OT}:
\begin{equation}\label{eq:11}
	\begin{matrix}
		\mathop{\max}\limits_{\bm{T}\in\mathbb{R}^{S\times k}_{+}}\texttt{Tr}(\bm{T}^{\top}\mathcal{P}^{\top}\bm{Z})+\kappa h(\bm{T}),\\
		s.t.\ \ \bm{T}^{\top}\bm{1}_{k}=\bm{1}_{S},\ \bm{T}\bm{1}_{S}=\frac{S}{k}\bm{1}_{k},
	\end{matrix}
\end{equation}
where $h(\bm{T})$ is an entropy, and $\kappa>0$ is a parameter that controls the smoothness of the distribution. The efficient solver on GPU of Eq. \ref{eq:11} can be given as the Sinkhorn algorithm \cite{GPUOT1, GPUOT2}. Our online subspace clustering involves few matrix multiplications, so it is computed by few steps of iteration.

With Eq. \ref{eq:6}, the prototype $\mathcal{P}_{j}$ is estimated from the pixel-wise feature embeddings that are with the highest confidence of clusters $j$ ($j=1,\cdots,k$). Specifically, for all pixels assigned to subspace $j$, the prototype $\mathcal{P}_{j}$ can be derived as the average of the pixel-wise embeddings, which is the center of the pixel embedding within segment $j$. 

The proposed online subspace clustering method is performed as follows: pixels with the same motion pattern are first assigned to the prototype $\mathcal{P}_{j}$ belonging to that subspace $\mathcal{S}_{j}$, and then the prototypes are updated according to the assignments. It is natural to derive a training objective for pixel assignment from Eqs. \ref{eq:5} and \ref{eq:7} as
\begin{equation}\label{eq:12}
	\mathcal{L}_{pc}=\frac{1}{S}\sum_{s=1}^{S}(1-{\mathbf{z}^{(s)}}^{\top}\mathcal{P}_{\alpha^{(s)}})^{2},
\end{equation}
where $\mathcal{L}_{pc}$ indicates the prototypes' discrinimitiveness, and
\begin{equation}\label{eq:13}
	\mathcal{L}_{cc}=-\frac{1}{S}\sum_{s=1}^{S}\log\frac{\exp({\mathbf{z}^{(s)}}^{\top}\mathcal{P}_{\alpha^{(s)}})}{\exp(\sum_{j=1}^{k}{\mathbf{z}^{(s)}}^{\top}\mathcal{P}_{j})},
\end{equation}
$\mathcal{L}_{cc}$ is the cluster contrastive loss.

\subsection{Contrastive Learning based on a Boundary Prior}

Considering the fact that the motion of the foreground object is different from that of the background \cite{SP}, a pixel-level contrastive learning strategy is introduced to improve the feature discrimination between the foreground and background. A core design philosophy for this strategy is that the average motion $\mathbf{m}$ of the boundary pixel is compared with all pixels $\mathbf{x}^{(s)}$ ($s\in\mathbb{R}^{H\times W}$) in the optical flow frame, so as to judge the similarity between the pixels and the boundary motion to determine whether it belongs to the background region. The cosine similarity between each pixel $\mathbf{x}^{(s)}$ and $\mathbf{m}$ is first computed as:
\begin{equation}\label{eq:14}
	s^{(s)}=1-\frac{{\mathbf{x}^{(s)}}^{\top}\mathbf{m}}{\parallel\mathbf{x}^{(s)}\parallel\parallel\mathbf{m}\parallel}.
\end{equation}

To determine the background region, a threshold $\delta$ is set to binarize the similarity map derived by Eq. \ref{eq:14}. After then, the foreground and background sets are obtained by $\mathcal{K}^{+}=\{\mathbf{x^{(+)}}|s^{(s)}<\delta\}$ and $\mathcal{K}^{-}=\{\mathbf{x^{(-)}}|s^{(s)}\geqslant\delta\}$, respectively. Thus, the foreground region $\mathcal{K}^{+}$ and the average motion $\mathbf{m}_{f}$ of the foreground pixels are treated as a pair. And the same is true for the background region $\mathcal{K}^{-}$ and the average motion $\mathbf{m}_{b}$ of the background pixels. Our contrastive learning framework aims to maximize the distance between the foreground and background representations. The final saliency contrastive loss is formulated as:
\begin{equation}\label{eq:15}
	\begin{matrix}
		&&\mathcal{L}_{sc}=&\\
		&&-\frac{1}{\parallel\mathcal{K}^{-}\parallel}\sum_{\mathcal{K}^{-}}\log\frac{\exp({\mathbf{x}^{(-)}}^{\top}\mathbf{m}_{b})}{\exp({\mathbf{x}^{(-)}}^{\top}\mathbf{m}_{b})+\exp({\mathbf{x}^{(-)}}^{\top}\mathbf{m}_{f})}&\\
		&&-\frac{1}{\parallel\mathcal{K}^{+}\parallel}\sum_{\mathcal{K}^{+}}\log\frac{\exp({\mathbf{x}^{(+)}}^{\top}\mathbf{m}_{f})}{\exp({\mathbf{x}^{(+)}}^{\top}\mathbf{m}_{f})+\exp({\mathbf{x}^{(+)}}^{\top}\mathbf{m}_{b})}.&
	\end{matrix}
\end{equation}
When the contrastive loss $\mathcal{L}_{sc}$ is applied to pull close and push apart the representations in positive and negative pairs, the motion pattern of the foreground object and the background in the optical flow are gradually separated.

\subsection{Optimization}

Stochastic gradient descent (SGD) is adopted to learn the parameters of the model, which consists of an auto-encoder, a feature MLP, and a spatial attention module. 

Now we show how to solve the VOS using our proposed contrastive motion clustering algorithm. We initialize the parameters of the auto-encoder by Eq. \ref{eq:3}. The prototypes $\mathcal{P}_{j}$ ($j=1,\cdots,k$) are initialized by a Gaussian distribution and are normalized to have unit $\ell_{2}$ norm.

At iteration $t$, we first obtain the embedding features $\bm{Z}$ using the auto-encoder and apply $\ell_{2}$ normalization. Then, we assign the label $\alpha$ to each pixel with a posterior probability as in Eq. \ref{eq:8}. In the cluster setting, we use the Sinkhorn algorithm \cite{GPUOT1} with hard assignment to group each pixel for the prototypes $\mathcal{P}_{j}$ ($j=1,\cdots,k$), as proposed in Eq. \ref{eq:11}. As the M-step of the Expectation-Maximization (EM) framework, the prototypes are then updated by accounting for the online clustering results. The non-learnable prototypes $\mathcal{P}_{j}$ are not learned by SGD, but are computed as the centers of the corresponding feature representations $\bm{Z}$. In particular, in each training iteration, each prototype is updated as:
\begin{equation}\label{eq:16}
	\mathcal{P}_{j} = \frac{1}{\lvert\mathcal{S}_{j} \rvert}\sum_{\alpha^{(s)}=j}{\mathbf{z}^{(s)}},
\end{equation}
where $\lvert\mathcal{S}_{j}\rvert$ is the number of pixels belonging to this subspace $\mathcal{S}_{j}$, and $s$ denotes the spatial position. Meanwhile, we compute and binarize the saliency map by Eq. \ref{eq:14} to obtain $\mathcal{K}^{+}$ and $\mathcal{K}^{-}$ sets for boundary prior-based contrastive learning. The parameters of our model are directly optimized by minimizing the combinatorial loss over all training pixel samples from the each video:
\begin{equation}\label{eq:17}
	\mathcal{L}=\mathcal{L}_{c}+\lambda_{1}\mathcal{L}_{pc}+\lambda_{2}\mathcal{L}_{cc}+\lambda_{3}\mathcal{L}_{sc}.
\end{equation}
After performing each training iteration, the cluster labels are obtained from the maximum matching formula in Eq. \ref{eq:6}. To guide the object segmentation, we use the boundary motion information as a prior. We found the optimal assignment between foreground and background with the Hungarian algorithm, using the cosine similarity between each prototype $\mathcal{P}_{j}$ and the background region $\mathcal{K}^{-}$ as a cost function with a threshold of $\eta$. Unlike the foreground region $\mathcal{K}^{+}$, the background region $\mathcal{K}^{-}$ tends to be more robust to noise than it.

For each frame, the online subspace clustering is then performed to achieve unsupervised motion segmentation. The whole optimization process is detailed in Algorithm \ref{alg:algorithm1}. The proposed method achieves a joint optimization of subspace clustering and embedded representation learning.

\begin{algorithm}[t]
	\caption{Contrastive clustering of optical flow for online UVOS}
	\label{alg:algorithm1}
	\KwIn{Optical flow; Number of clusters $k$; Embedding dimension $r$; Subspace dimension $p$; Hyperparameters $\kappa$, $\delta$, $\eta$, $\lambda_{1}$, $\lambda_{2}$ and $\lambda_{3}$; Maximum iteration $T_{\max}$.}
	\KwOut{Cluster labels of pixels.}  
	\BlankLine
	Initialize auto-encoder by minimizing Eq. \ref{eq:3}.
	\BlankLine
	Initialize non-learnable prototypes from Gaussian distribution and apply $\ell_{2}$ normalization.
	
	\While{$t\leqslant T_{\max}$}{
		Learn embedded representation $\bm{Z}$.
		
		Compute cluster labels by solving Eqs. \ref{eq:8} and \ref{eq:11}.
		
		Update non-learnable prototypes according to cluster labels as in Eq. \ref{eq:16}.
		
		Compute and binarize the saliency map by Eq. \ref{eq:14} to obtain $\mathcal{K}^{+}$ and $\mathcal{K}^{-}$ sets, respectively.
		
		Update the network parameters by minimizing the objective function in Eq. \ref{eq:17}.
		}
	
	Use Eq. \ref{eq:6} to obtain the final updated cluster labels.
	
	Assign pixel-wise foreground/background labels with the Hungarian algorithm based on boundary prior.
	
	\KwRet{Foreground/Background labels of pixels.}
\end{algorithm}

\section{Experiments}
\label{sec:exp}

\subsection{Experimental Setup}\label{Exp:ExpSet}

\noindent\textbf{Datasets and evaluation metrics}. To test the performance of our online subspace clustering, we carry out comprehensive experiments on the following three UVOS datasets:

\DAVIS\ \cite{DAVIS2016} is currently the most popular VOS benchmark, consisting of 50 high-quality video sequences (30 videos for the \texttt{train} set and 20 for the \texttt{val} set). Each frame is densely annotated with pixel-wise ground truth for the foreground objects. We perform online clustering and evaluation on the validation set. For quantitative evaluation, we adopt standard metrics suggested by \cite{DAVIS2016}, namely region similarity $\mathcal{J}$, which is the intersection-over-union of the prediction and ground-truth, computing the mean over the \texttt{val} set.

\FBMS \cite{FBMS} contains videos of multiple moving objects, providing test cases for multiple object segmentation. The \FBMS\ has 59 sparsely annotated video sequences, with 30 sequences for validation.

\STv\ \cite{SegTrackV2} contains 14 densely annotated videos and 976 annotated frames. Each sequence contains 1-6 moving objects and presents challenges such as motion blur, appearance change, complex deformation, occlusion, slow motion, and interacting objects.

Following the evaluation protocol in \cite{CIS}, we combine multiple objects as a single foreground and use the region similarity $\mathcal{J}$ to measure the segmentation performance for the \FBMS\ and \STv.

\noindent\textbf{Implementation details.} The optical flow is estimated by using the RAFT \cite{RAFT} and FlowFormer \cite{FlowFormer}. The flows are resized to the size of the original image \cite{MG}, with each input frame having a size of 480$\times$854 for the \DAVIS\ and 480$\times$640 for the \FBMS\ and \STv. We convert the optical flow to 3-channel images with the standard visualization used for the optical flow and normalize it to [-1, 1], and use only the previous frames for the optical flow estimation in the online setting.

We construct our model with a CNN encoder of architecture [64, MP, 128, MP, 256] and a decoder with deconvolutional layers (or transposed convolution) \cite{DCNN1, DCNN2} that can be used for learnable guided upsampling of intermediate encoder representations. Here, MP denotes a max pooling layer with stride 2. The output dimension of the embedding network $\Phi(\cdot)$ is 256, \emph{i.e.} $r$=256. In MLP, the number of hidden units is [256, 256] with ReLU as the activation function for the hidden layer. The output dimension $p$ is 10. We first initialize our auto-encoder by pre-training 10 epochs on \DAVIS\ \texttt{val}, which takes about 7 minutes for \DAVIS\ with 480$\times$854 resolution. The whole network is trained using the Adam \cite{Adam} optimizer ($\beta_{1}$=0.9 and $\beta_{2}$=0.999) with a learning rate of $10^{-3}$. The hyper-parameters are set empirically to: $\kappa$=0.05, $\delta$=0.1, $\eta$=0.5, $\lambda_{1}$=$\lambda_{2}$=$\lambda_{3}$=0.01, and $T_{\max}$=100. We also discuss the impact of different values of the hyper-parameters in Section \ref{Exp:Abla}.

\subsection{Ablation Studies}\label{Exp:Abla}

\begin{table}[t]
	\centering
	\captionsetup{font=small}
	\caption{Comparison of the three different optical flow methods as the input on the \DAVIS\ dataset, measured by the mean $\mathcal{J}$. In the inference step, we employ multi-scale and CRF to improve the final performance of MG \cite{MG}.}
	\label{tab:Abla:op}
	\vspace{-3pt}
	\resizebox{0.7\columnwidth}{!}{
		\tablestyle{12pt}{1.02}
		\begin{tabular}{@{}r|c|c@{}}
			\toprule 
			Flow~~~~~ & Method & Mean $\mathcal{J}\uparrow$ \\ 
			\midrule
			\multirow{2}[0]{*}{~~PWC-Net \cite{PWC}} & MG \cite{MG}    & 63.7 \\
			& \textbf{Ours}  &  67.9 \textcolor{tablegreen}{\scriptsize (\textbf{+4.2})}\\
			\multirow{2}[0]{*}{~~RAFT \cite{RAFT}} & MG \cite{MG}   & 68.3 \\
			& \textbf{Ours}  & 72.0 \textcolor{tablegreen}{\scriptsize (\textbf{+3.7})} \\
			\multirow{2}[0]{*}{~~FlowFormer \cite{FlowFormer}} & MG \cite{MG}   & 70.3 \\
			& \textbf{Ours}  & 75.4 \textcolor{tablegreen}{\scriptsize (\textbf{+5.1})} \\
			\bottomrule
		\end{tabular}
	}
\end{table}

To demonstrate the influence of each component and hyper-parameters in our method, we perform an ablation study on the \DAVIS\ \texttt{val} set. The evaluation criterion is the mean region similarity ($\mathcal{J}$).

\noindent\textbf{Choice of optical flow algorithm.} Our model takes only the optical flow as the input to solve the motion grouping problem. Table \ref{tab:Abla:op} shows the effect of the quality of different inputs. With the same optical flow estimation methods (\emph{i.e.}, PWC-Net \cite{PWC}, RAFT \cite{RAFT}, and FlowFormer \cite{FlowFormer}), our proposed algorithm outperforms MG \cite{MG} by \textbf{4.2}\%, \textbf{3.7}\%, and \textbf{5.1}\% points, respectively, in terms of mean $\mathcal{J}$ on the \DAVIS\ \texttt{val} set. The improved optical flow model (FlowFormer) further enlarges performance gains. Thus, the optical flow estimated by the FlowFormer is the input to our model.

\noindent\textbf{Effectiveness of spatial attention module.} To verify the effect of the spatial attention module $A(\cdot)$ in Eq. \ref{eq:2}, we gradually remove the spatial attention module $A(\cdot)$, the average pooling, and max pooling in our auto-encoder, denoted as AE, \emph{w/.} max pooling, and \emph{w/.} average pooling, respectively. This means that the embedding features $\bm{Z}$ are directly fed into the decoder $\Psi(\cdot)$, where we employ AE as the baseline for the ablation study. The results can be referred to in Table \ref{tab:Abla:variant}. Compared to the baseline, the variants with max pooling and average pooling can independently boost the performance by \textbf{2.4}\% and \textbf{2.2}\%, respectively. Based on these high-performance variants, the spatial attention module $A(\cdot)$, which combines max and average pooling operations to enhance losing important information on the regions of object boundaries, further improves the performance by \textbf{3.3}\% in terms of mean $\mathcal{J}$. This demonstrates the superiority of the spatial attention module.

\begin{table}[t]
	\centering
	\captionsetup{font=small}
	\caption{Ablation study of the spatial attention module on the \DAVIS\ dataset, measured by the mean $\mathcal{J}$. We employ an auto-encoder, which is implemented by directly connecting the encoder $\Phi(\cdot)$ and the decoder $\Psi(\cdot)$ as the baseline for all experiments, denoted as AE.}
	\label{tab:Abla:variant}
	\vspace{-3pt}
	\resizebox{0.6\columnwidth}{!}{
		\tablestyle{12pt}{1.02}
		\begin{tabular}{@{}l|c@{}}
			\toprule	~~~~Network Variant & Mean $\mathcal{J}\uparrow$ \\ 
			\midrule
			AE (baseline) 					& 72.1 	\\
			AE \emph{w/.} max pooling 		& 74.5 \textcolor{tablegreen}{\scriptsize (\textbf{+2.4})}	\\
			AE \emph{w/.} average pooling 	& 74.3 \textcolor{tablegreen}{\scriptsize (\textbf{+2.2})}	\\
			AE \emph{w/.} $A(\cdot)$		& 75.4 \textcolor{tablegreen}{\scriptsize (\textbf{+3.3})}	\\
			\bottomrule
		\end{tabular}
	}
\end{table}

\begin{figure*}[t]
	\centering
	\begin{tabular}{@{}c@{\hskip 1pt}c@{\hskip 1pt}c@{\hskip 1pt}c@{\hskip 1pt}c@{\hskip 1pt}c@{}}
		\includegraphics[width=0.16\linewidth]{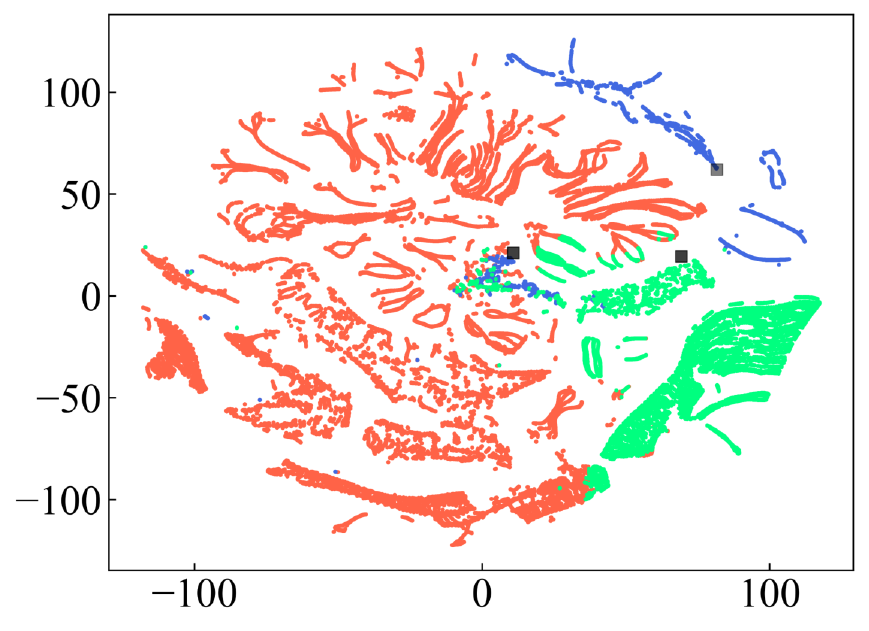} 		&
		\includegraphics[width=0.16\linewidth]{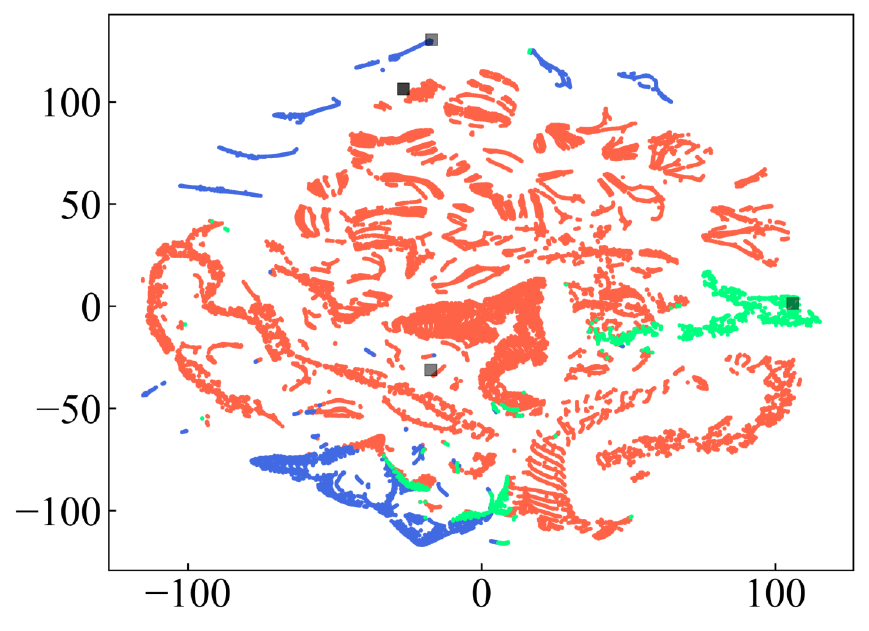}		&
		\includegraphics[width=0.16\linewidth]{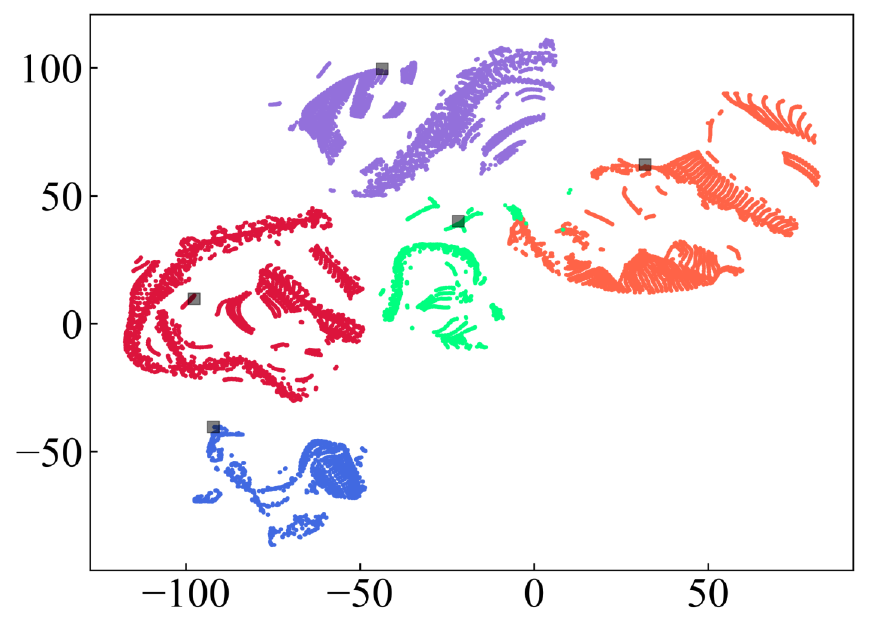}	&
		\includegraphics[width=0.16\linewidth]{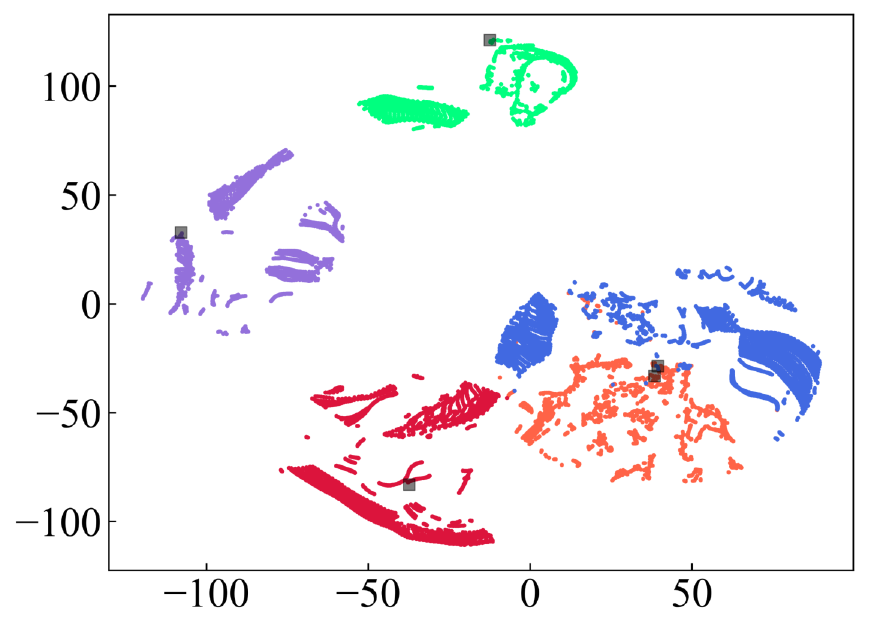}	&
		\includegraphics[width=0.16\linewidth]{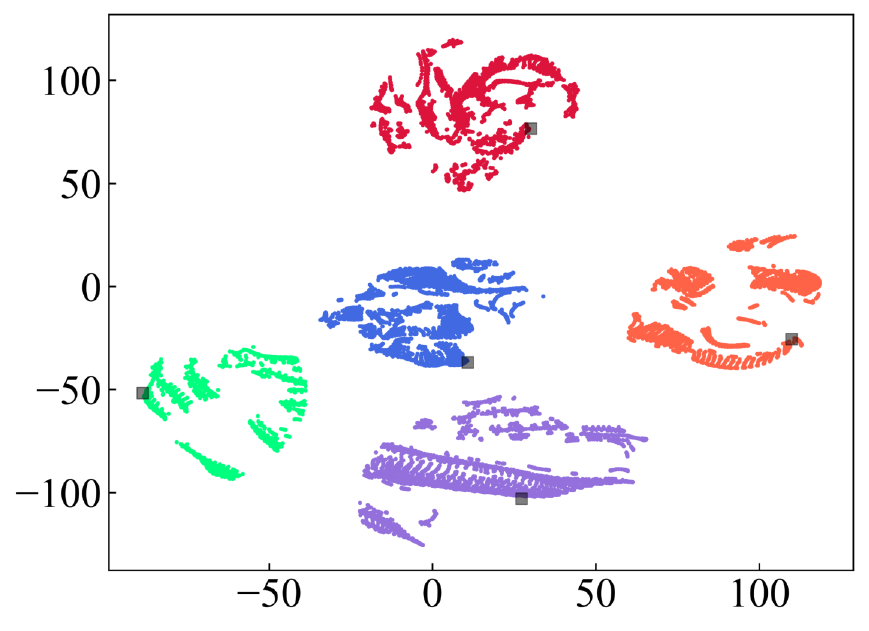}	& 
		\includegraphics[width=0.16\linewidth]{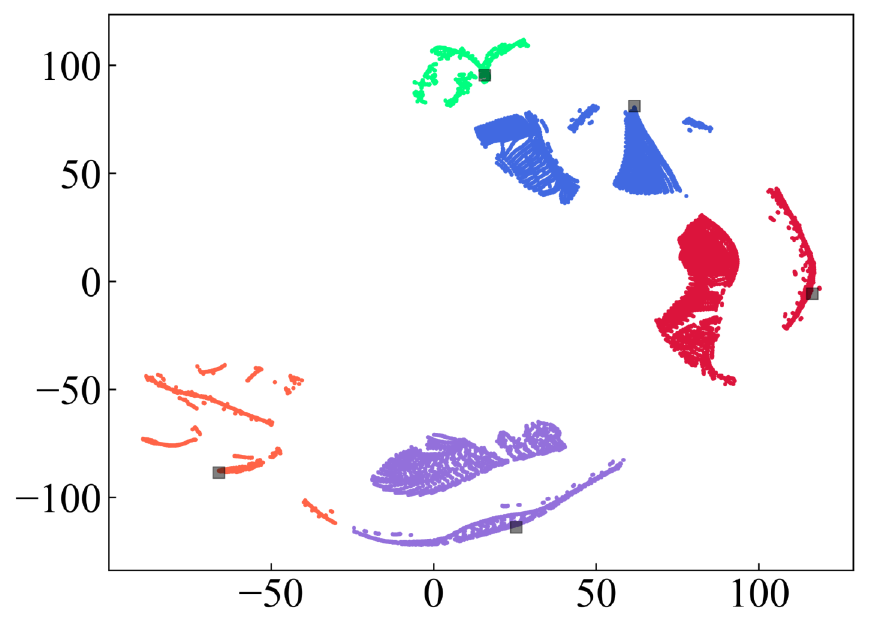}	\\
		~~~Vector $\bm{0}$ 							&
		~~~Vector $\bm{1}$ 							&
		~~~Orthogonal      							&
		~~~~$\mathcal{U}(0,1)$						&
		~~~~$\mathcal{N}(0,1)$						&
		~~~Truncated $\mathcal{N}(0,1)$
	\end{tabular}
	\vspace{-3pt}
	\caption{Visualization of the embedded representations $\bm{Z}$ with t-SNE \cite{tSNE} on the \textit{bmx-trees} sequence from the \DAVIS\ dataset. Note that the number of prototypes $k$ is set to 5 for each initialization condition, and we optimize our model for 10 iterations on each frame. \tikzsymbol[rectangle]{fill=black,minimum width=4pt}{2pt} represents the each prototype $\mathcal{P}_{j}$.}
	\label{fig:tSNE-init}
\end{figure*}

\begin{figure*}[!ht]
	\centering
	\begin{tabular}{@{}c@{\hskip 1pt}c@{\hskip 1pt}c@{}}
		\begin{overpic}[width=0.33\linewidth]{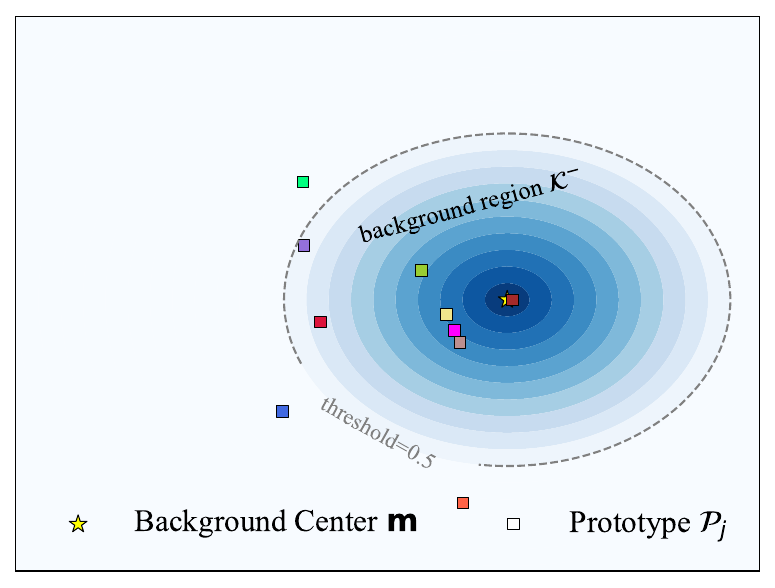}
			\put(3, 54){\includegraphics[width=0.11\linewidth]{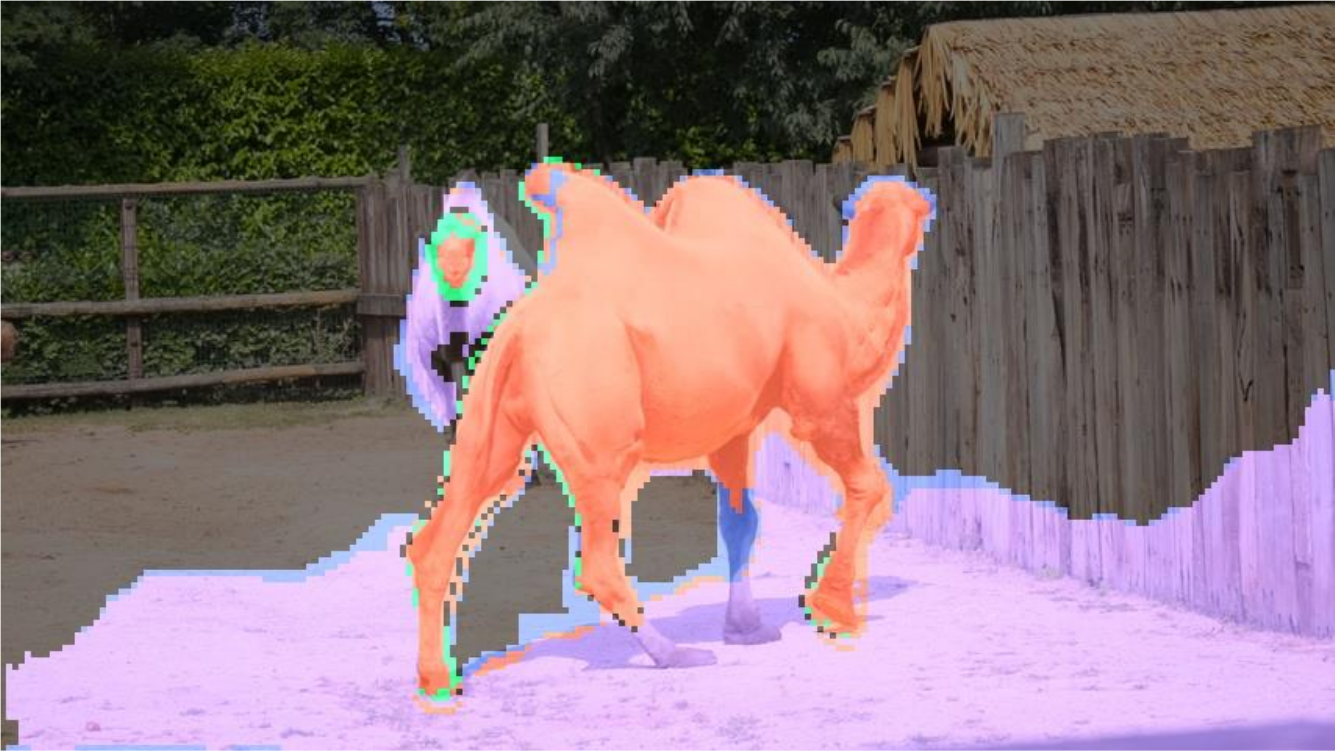}}
		\end{overpic}	&
		\begin{overpic}[width=0.33\linewidth]{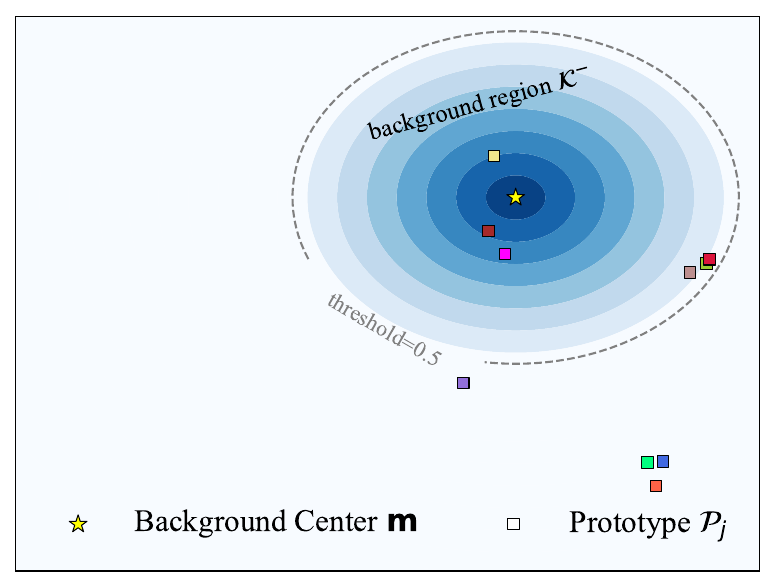}
			\put(3, 54){\includegraphics[width=0.11\linewidth]{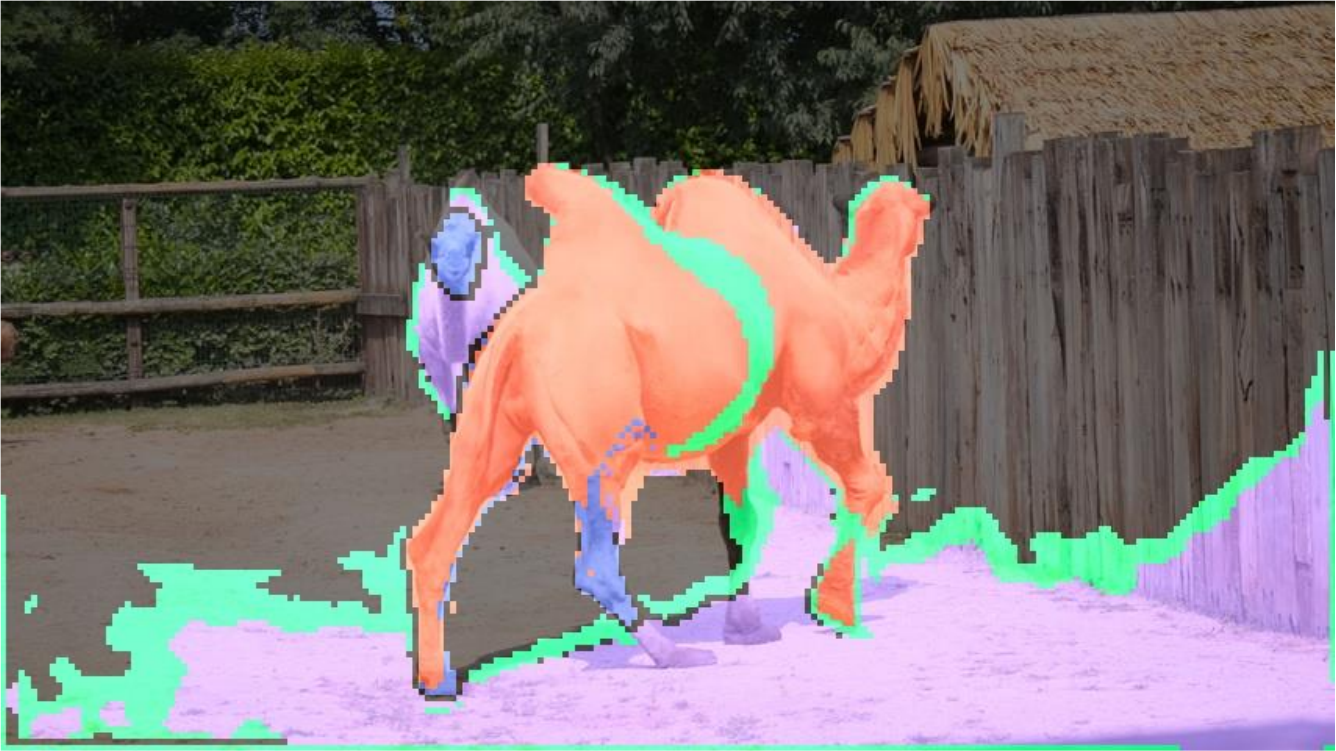}}
		\end{overpic}	&
		\begin{overpic}[width=0.33\linewidth]{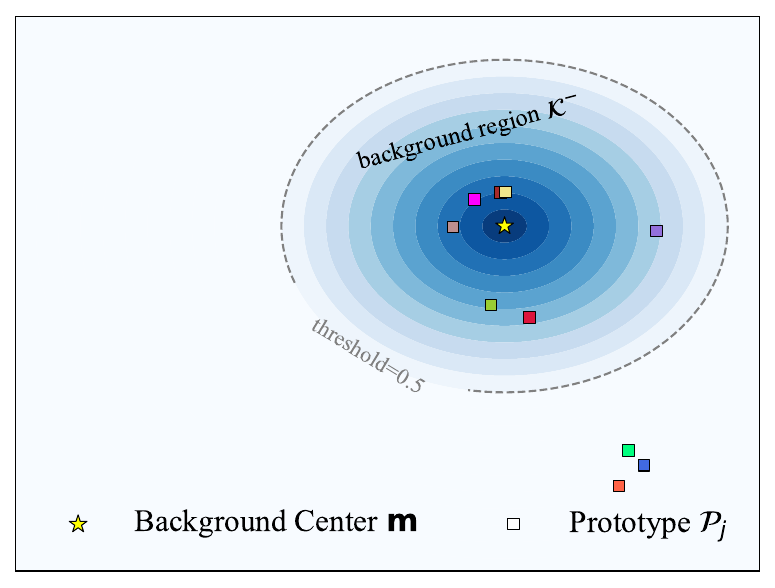}
			\put(3, 54){\includegraphics[width=0.11\linewidth]{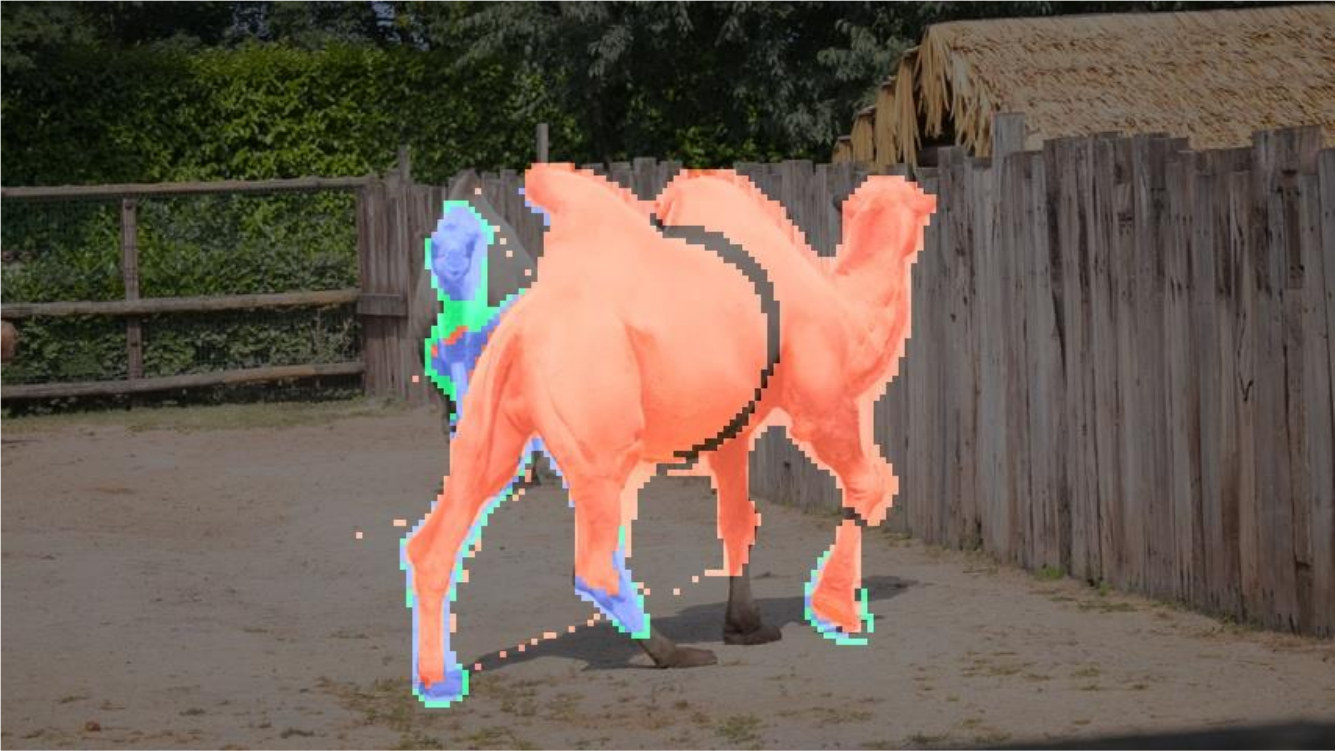}}
		\end{overpic}	\\
		\vspace{5pt}
		Iteration 0 														&
		Iteration 20 														&
		Iteration 40   														\\
		\begin{overpic}[width=0.33\linewidth]{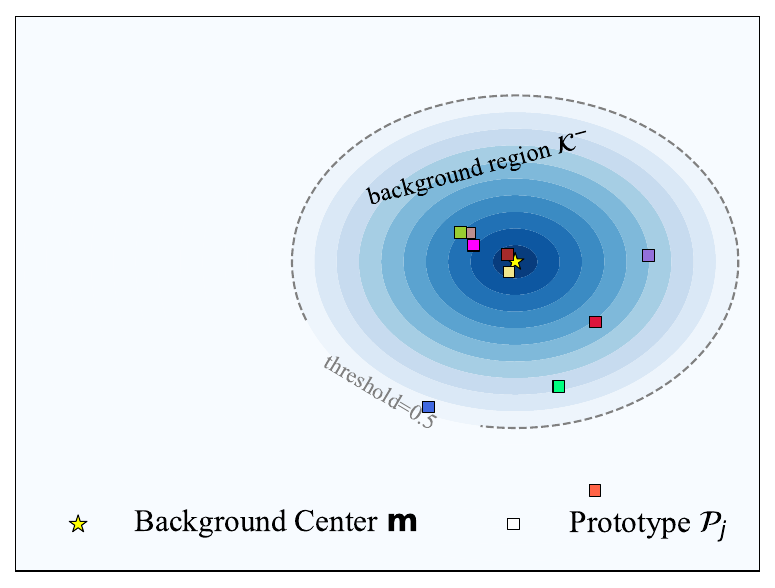}
			\put(3, 54){\includegraphics[width=0.11\linewidth]{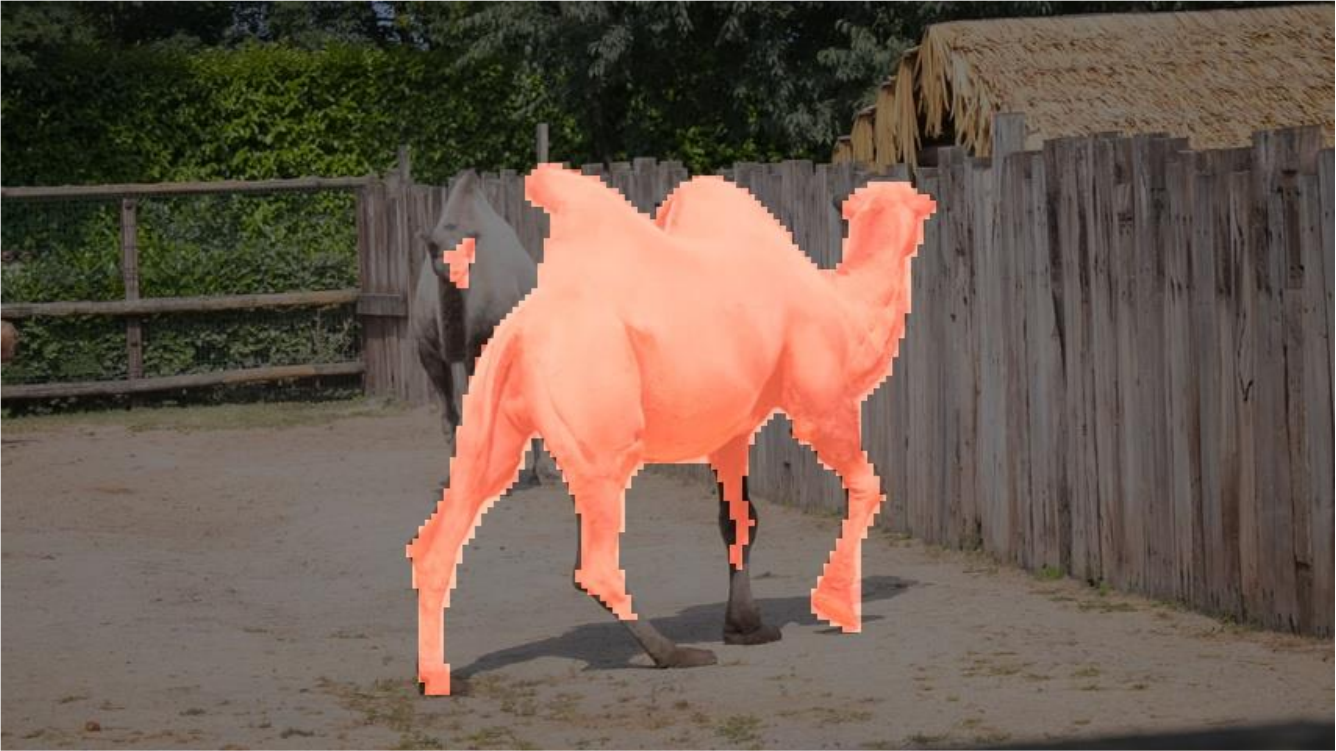}}
		\end{overpic}	&
		\begin{overpic}[width=0.33\linewidth]{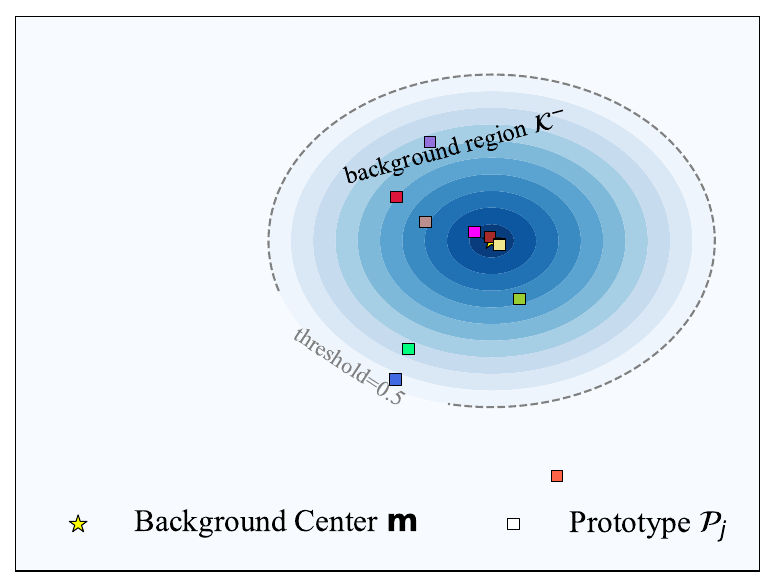}
			\put(3, 54){\includegraphics[width=0.11\linewidth]{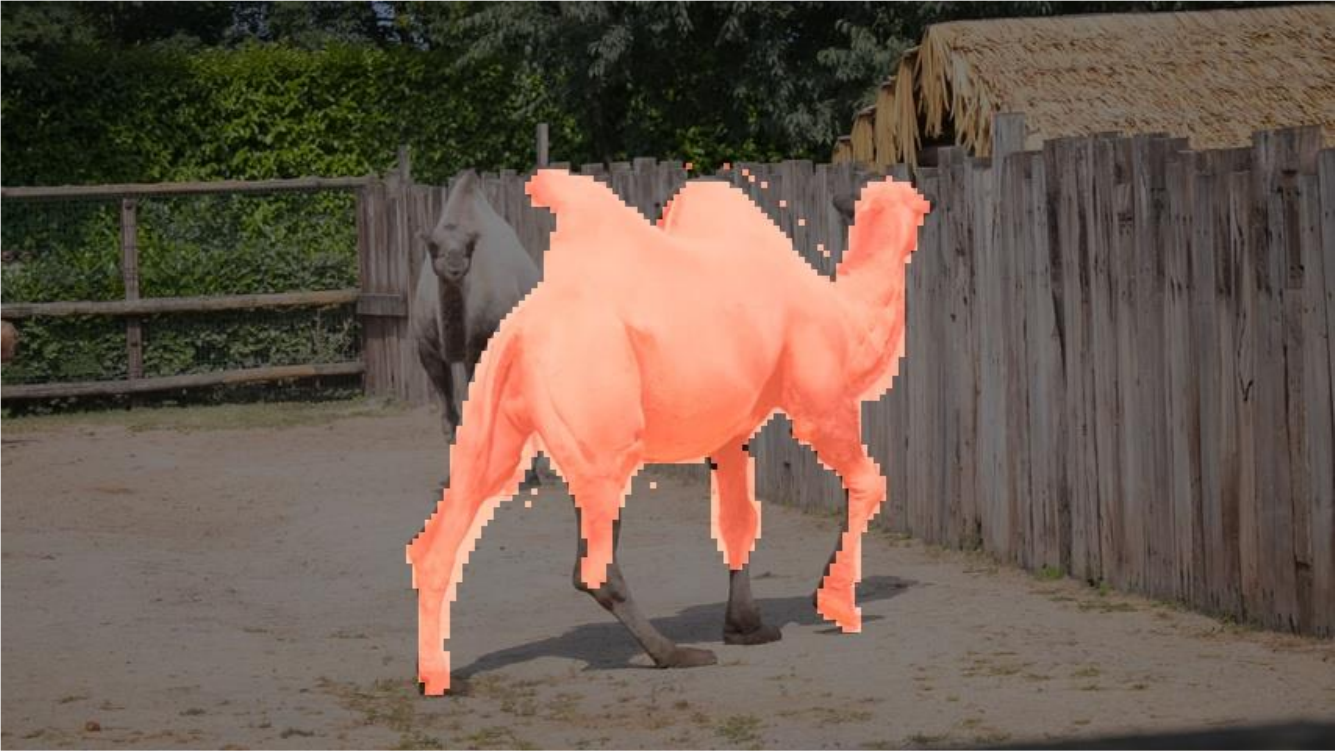}}
		\end{overpic}	& 
		\begin{overpic}[width=0.33\linewidth]{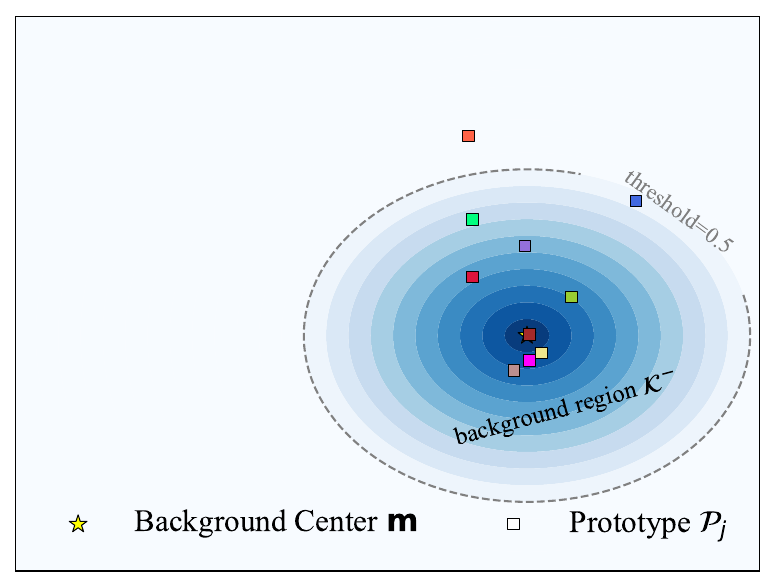}
			\put(3, 54){\includegraphics[width=0.11\linewidth]{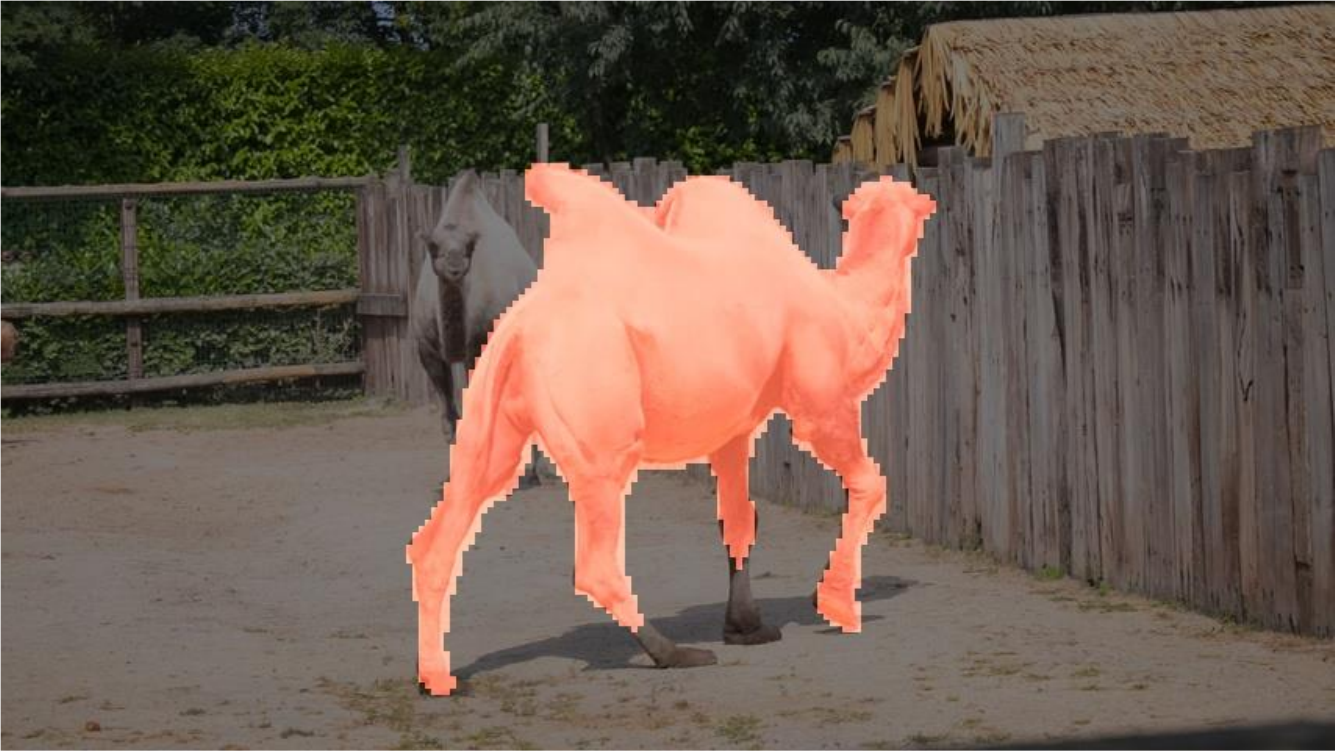}}
		\end{overpic}	\\
		Iteration 60														&
		Iteration 80														&
		Iteration 100														
	\end{tabular}
	\vspace{-3pt}
	\caption{Visualization of the distribution of the background region $\mathcal{K}^{-}$ (blue region) and the each prototype $\mathcal{P}_{j}$ (square) during the training iteration. Based on the cosine similarity between each prototype $\mathcal{P}_{j}$ and the background region $\mathcal{K}^{-}$, we draw a contour with a threshold of 0.5. A darker color indicates a higher similarity. Based on the observations, the distribution of prototypes is iteratively refined. The prototype of the foreground objects (\tikzsymbol[rectangle]{fill=tomato,minimum width=4pt}{2pt}) far away from that of the background center (\tikzsymbol[star]{fill=yellow,star point ratio=2.618}{1pt}) and the background distractors (\emph{e.g.}, \tikzsymbol[rectangle]{fill=royalblue,minimum width=4pt}{2pt} and \tikzsymbol[rectangle]{fill=springgreen,minimum width=4pt}{2pt}) can be filtered by contrastive learning based on a boundary prior. The segmentation results of each iteration are shown in the upper left corner of each figure. t-SNE \cite{tSNE} is used to reduce the dimensionality of the features.}
	\label{fig:tSNE-camel}
\end{figure*}

\begin{table*}[h]
	\captionsetup{font=small}
	\caption{Ablation studies of the proposed method on the \DAVIS\ dataset, measured by the mean $\mathcal{J}$.}
	\label{tab:Abla}
	\vspace{-3pt}
	\subfloat[{Training Objective $\mathcal{L}$} \label{table:Abla:loss}]{
		\tablestyle{8pt}{1.05}
		\begin{tabular}{@{}cccc|c@{}}
			\toprule
			$\mathcal{L}_{c}$     & $\mathcal{L}_{sc}$     &  $\mathcal{L}_{pc}$    &   $\mathcal{L}_{cc}$   & Mean $\mathcal{J}\uparrow$ \\ \midrule 
			\cmark & 		&		 &  	  & 73.9	\\
			\cmark & \cmark &		 &   	  &	74.4 \textcolor{tablegreen}{\scriptsize (\textbf{+0.5})}    \\
			\cmark & 		& \cmark &   	  &	74.3 \textcolor{tablegreen}{\scriptsize (\textbf{+0.4})}	\\
			\cmark & 		& 		 & \cmark &	74.1 \textcolor{tablegreen}{\scriptsize (\textbf{+0.2})}	\\
			\cmark & 		& \cmark & \cmark &	74.7 \textcolor{tablegreen}{\scriptsize (\textbf{+0.8})}	\\
			\cmark & \cmark & \cmark & \cmark & \textbf{75.4} \textcolor{tablegreen}{\scriptsize (\textbf{+1.5})} 	\\
			\bottomrule
		\end{tabular}
}\hfill
\subfloat[{Initialization of prototypes}\label{table:Abla:init}]{%
		\tablestyle{10pt}{1.05}
		\begin{tabular}{@{}c|c@{}}
			\toprule
			Init. Method & Mean $\mathcal{J}\uparrow$\\ \midrule
			Vector $\bm{0}$  	& 44.4 	\\
			Vector $\bm{1}$ 	& 43.5 	\\
			Orthogonal \cite{Orthogonal}  	& 75.1 	\\
			$\mathcal{U}(0,1)$  	& 74.8	\\
			$\mathcal{N}(0,1)$  	& \textbf{75.4}	\\
			Truncated $\mathcal{N}(0,1)$ & 75.3	\\
			\bottomrule
		\end{tabular}
}\hfill
\subfloat[{Number of clusters $k$}\label{table:Abla:nk}]{%
		\tablestyle{10pt}{1.05}
		\begin{tabular}{@{}l|c@{}}
			\toprule
			Clusters $k$ & Mean $\mathcal{J}\uparrow$ \\\midrule
			~~~$k$=2 & 65.6 \\
			~~~$k$=5 & 69.2 \\
			~~~$k$=10 & 73.9 \\
			~~~$k$=20 & 75.2 \\
			~~~$k$=30 & \textbf{75.4} \\
			~~~$k$=50 & 74.9 \\
			\bottomrule
		\end{tabular}
}\hfill
\subfloat[{Background threshold $\delta$}\label{table:Abla:delta}]{%
	\tablestyle{10pt}{1.05}
		\begin{tabular}{@{}l|c@{}}
			\toprule
			Threshold $\delta$ & Mean $\mathcal{J}\uparrow$ \\ \midrule
			~~~$\delta$=0.05 & 74.4\\
			~~~$\delta$=0.1 & \textbf{75.4} \\
			~~~$\delta$=0.2  & 74.6 \\
			~~~$\delta$=0.5  & 73.7 \\
			~~~$\delta$=0.7  & 73.2 \\
			~~~$\delta$=0.9  & 72.9 \\
			\bottomrule
		\end{tabular}
}\hfill
\end{table*}

\noindent\textbf{Training objective.} We investigate our overall training objective (Eq. \ref{eq:17}). As shown in Table \ref{table:Abla:loss}, the model with $\mathcal{L}_{c}$ alone achieves a mean $\mathcal{J}$ score of 73.9\%. Adding $\mathcal{L}_{sc}$ brings a gain (\emph{i.e.}, \textbf{0.5}\%), which shows that it effectively improves the discriminability of foreground and background. After applying $\mathcal{L}_{pc}$ or $\mathcal{L}_{cc}$ individually, we observe that our model achieves improvements (\emph{i.e.}, \textbf{0.4}\%/\textbf{0.2}\%), and their combinations further improve the performance by nearly \textbf{1.0}\%. These facts not only demonstrate the effectiveness of $\mathcal{L}_{pc}$ and $\mathcal{L}_{cc}$ but also indicate that the contributions of the two constraints are almost orthogonal. Finally, combining all the losses together leads to the best performance, yielding a mean $\mathcal{J}$ score of 75.4\%. This further confirms the effectiveness of our training objective.

\noindent\textbf{Initialization of prototypes.} We evaluate the different initialization strategies of the prototypes on the \DAVIS\ dataset to get a better impression of the performance. Table \ref{table:Abla:init} shows the results of prototypes initialized by the vector $\bm{0}$, the vector $\bm{1}$, the orthogonal vectors \cite{Orthogonal}, the uniform distribution $\mathcal{U}(0,1)$, the standard normal distribution $\mathcal{N}(0,1)$, and the truncated normal distribution $\mathcal{N}(0,1)$. We notice that the prototypes filled with vectors $\bm{0}$ and $\bm{1}$ yield the worst performance compared to initializing the prototypes randomly. An improper initialization of the prototypes is problematic, and the constant initialization cannot guarantee the orthogonality of the prototypes and prevent all of them from collapsing onto a single point. It reveals that our method is sensitive to the initialization. We also compare different random initializations for the prototypes. We see that the random initialization outperforms the constant and the Gaussian initialization outperforms all the strategies. Fig. \ref{fig:tSNE-init} presents the t-SNE \cite{tSNE} visualization of the learned embedded representation on the \textit{bmx-trees} sequence from the \DAVIS\ dataset. In particular, without the random initialization of the prototypes, the representation learned from the auto-encoder failed to find a good clustering structure, leading to a somewhat inferior visualization. In contrast, the embedded representation leaned from the model with random initialization becomes significantly discriminative, and the proposed method can achieve promising performance when we have a good initialization of prototypes.

\noindent\textbf{Number of clusters.} Table \ref{table:Abla:nk} reports the performance of our approach with regard to the number of clusters $k$. For $k$=2, we directly segment foreground-background into 2 groups. This baseline obtains a score of 65.6\%. We can see that as $k$ increases, the mean $\mathcal{J}$ first increases and then decreases. Furthermore, when we use more clusters (\emph{i.e.}, $k$=5), we see a clear performance boost (65.6\%$\rightarrow$69.2\%). The score improves further when $k$=20 or $k$=30 is allowed; however, increasing $k$ beyond 30 gives marginal returns in performance. Therefore, we empirically set $k$=30 for a better trade-off between the accuracy and computational cost.

\noindent\textbf{Background threshold.} To discriminate the foreground from the background distractors, we introduce boundary motion information as prior knowledge and propose a contrastive loss based on a boundary prior to guide object segmentation. Fig. \ref{fig:tSNE-camel} shows the t-SNE \cite{tSNE} visualization of the embedded representation learned by the proposed method on the camel sequence from the \DAVIS\ dataset in different iterations, as it is important to understand how the representation evolves during training. In this scenario, when a similar object ditractor and texture background appears (\emph{e.g.}, the small camel around the target object), our model fails to capture the primary target in early iterations. However, with the help of the contrastive loss based on a boundary prior, we see that the embedded representation of the foreground object becomes more and more discriminative as the training iterations increase. For each video, the background region $\mathcal{K}^{-}$ explicitly maintains the consistency of the motion across the entire video. We empirically choose $\delta$ for the foreground saliency map decision to evaluate our model. The results are listed in Table \ref{table:Abla:delta}. We can see that when as $\delta$ increases, the mean $\mathcal{J}$ decreases. As a result, we empirically set $\delta$=0.1 with the best performance.

\subsection{Comparison with SoTAs}\label{Exp:SoTA}

\begin{table*}[htbp]
	\captionsetup{font=small}
	\caption{Quantitative results on the \texttt{val} set of video object segmentation benchmarks, using the region similarity $\mathcal{J}$. The best performance scores are highlighted in \textbf{bold}. The Extra Model in the fifth column denotes the required pre-training model. Runtime excludes the optical flow computation. OF and RGB represent the optical flow and RGB image, respectively.}
	\label{tab:SoTA}%
	\vspace{-3pt}
	\begin{threeparttable}
		\begin{center}
			\resizebox{1.0\linewidth}{!}{
				\tablestyle{5pt}{1.02}
				\begin{tabular}{@{}r|c|cc|c|c|ccc|r@{}}\toprule
					\multirow{2}[0]{*}{Method~~} & Training \& & \multirow{2}[0]{*}{OF} & \multirow{2}[0]{*}{RGB} & Input & Extra & \multicolumn{3}{c|}{Mean $\mathcal{J}\uparrow$} & Runtime $\downarrow$\\
					& Optimization &       &       & Resolution & Model & \multicolumn{1}{c}{\DAVIS} & \multicolumn{1}{c}{\FBMS} & \multicolumn{1}{c|}{\STv} & (s/frames)  \\\midrule
					FSEG \cite{FSEG} &   \tablecircle    &   \cmark    &   \cmark   &   480$\times$854    &   -    & 71.6  &  - & 61.4  & 7.2~~~~ \\
					LMP \cite{LMP} &   \tablecircle    &   \cmark    &      &   480$\times$854    &   -    & 64.5  &  - & -  & 18.3~~~~ \\
					ELM \cite{ELM}   & \tablestar &   \cmark    &    \cmark   &    -   &   -    & 61.8  & 61.6  & -      & -~~~~~~ \\
					SAGE \cite{SAGE} &     \tablestar  &   \cmark    &       &    -   &    -   & 42.6  & 61.2  & 57.6  & 0.9~~~~ \\
					CVOS \cite{CVOS} &   \tablestar    &   \cmark    &      &   480$\times$854    &   -    & 51.4  &  - & -  & -~~~~~~ \\
					SFM \cite{SFM} &   \tablestar    &       &   \cmark   &   480$\times$854    &   -    & 53.2  &  - & -  & 0.15~~~~ \\
					UOVOS \cite{UOVOS} &   \tablestar    &   \cmark    &  \cmark    &   480$\times$854    &   Mask R-CNN \cite{MaskRCNN}   & 74.6  &  63.9 & 61.5  & 0.36~~~~ \\ \midrule
					\textbf{Ours}  &   \tablestar    &  \cmark     &       &   480$\times$854    &    -   & \textbf{75.4}  &    \textbf{66.8}   &   \textbf{62.6}    & \textbf{0.12}~~~~ \\\bottomrule
				\end{tabular}%
			}
		\end{center}
		\begin{tablenotes}
			\footnotesize
			\item[]\tablestar\ is fully unsupervised and the model is optimized in a single input frame. \tablecircle\ is pre-trained on a video dataset.
		\end{tablenotes}
	\end{threeparttable}
\end{table*}%

\begin{figure*}[t]
	\centering
	\begin{tabular}{@{}c@{\hskip 1pt}c@{\hskip 1pt}c@{}}
		\includegraphics[width=0.33\linewidth]{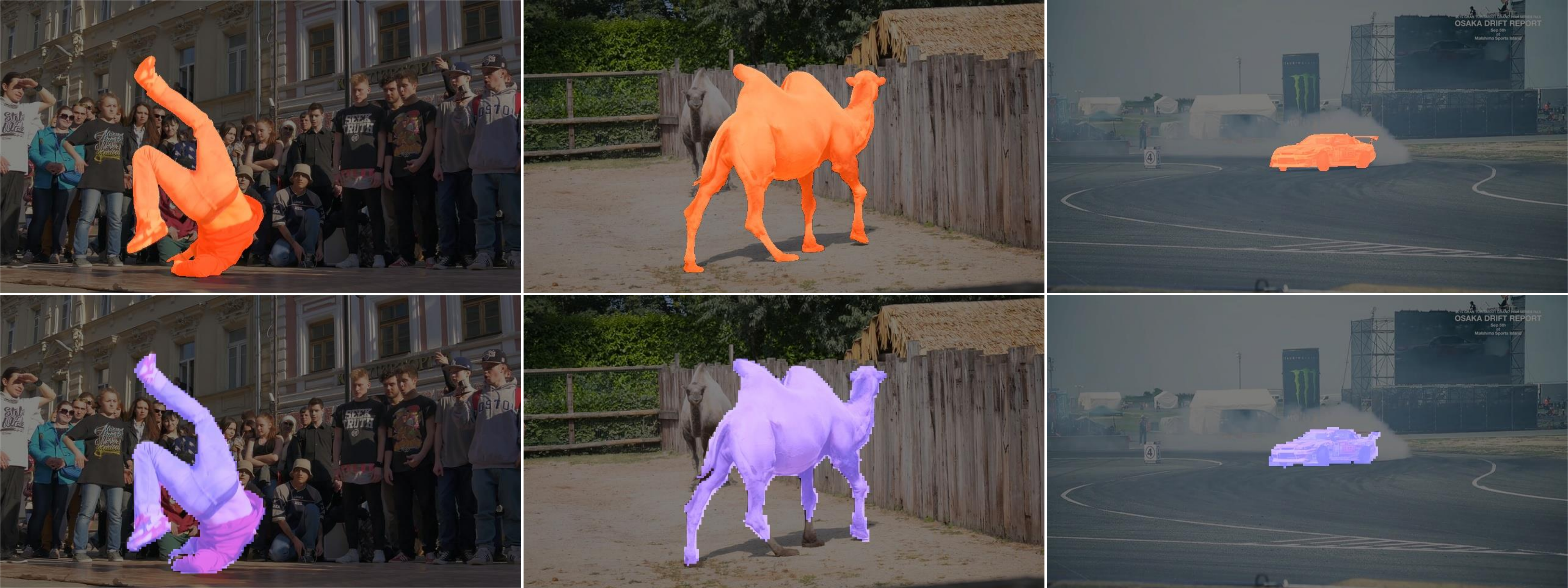} &
		\includegraphics[width=0.33\linewidth]{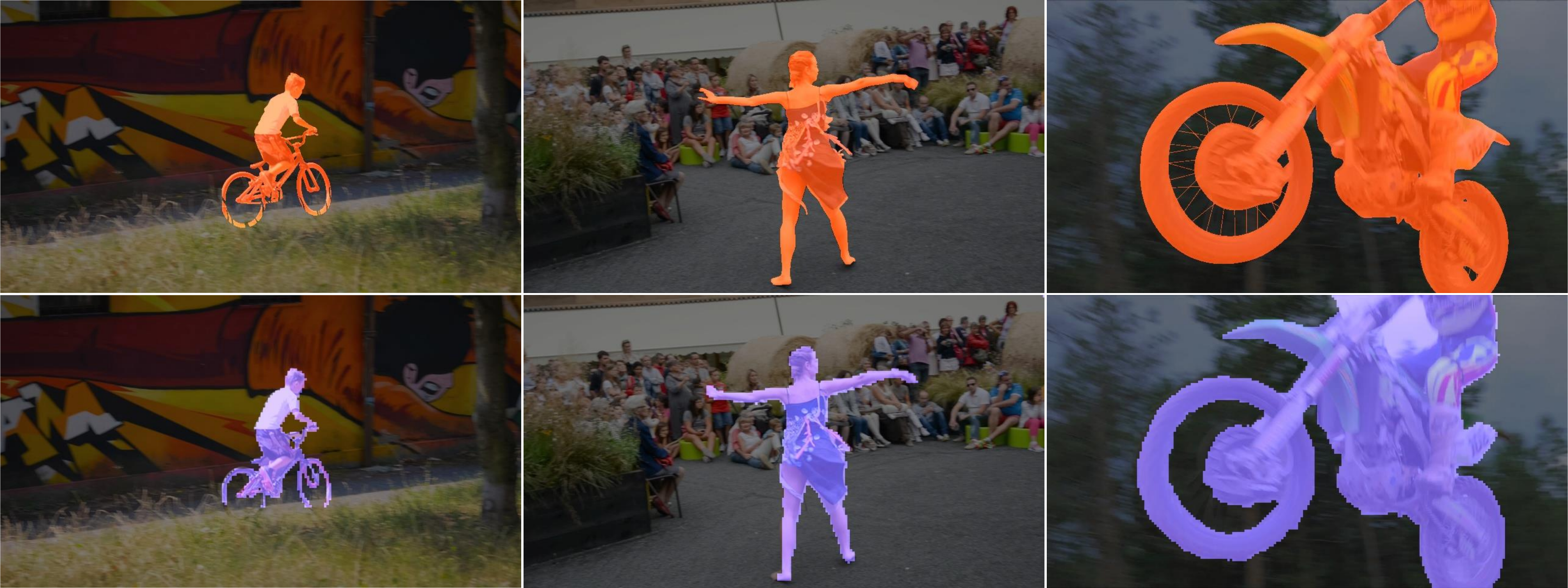} &
		\includegraphics[width=0.33\linewidth]{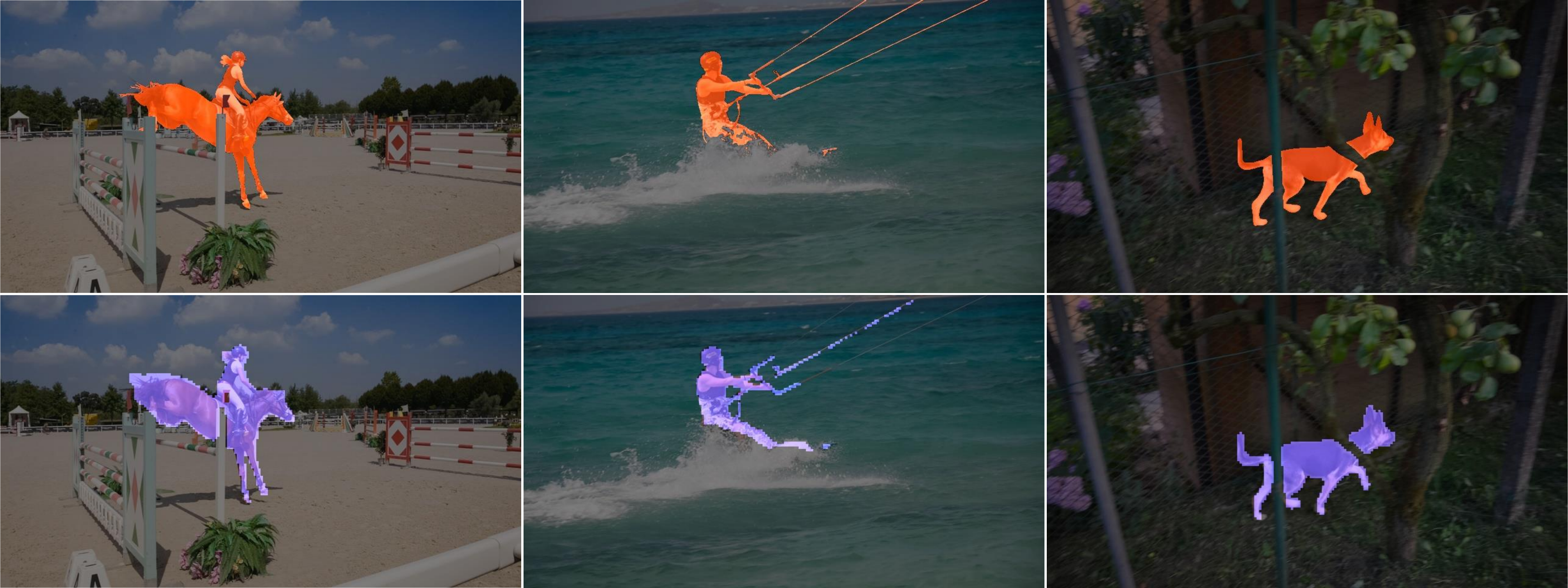} \\
		\vspace{5pt}
		Dynamic background &
		Motion blur &
		Occlusion			
	\end{tabular}
	\vspace{-3pt}
	\caption{Qualitative results of the proposed method on challenging scenarios from the \DAVIS. From left to right: dynamic background (\textit{breakdance}, \textit{camel}, and \textit{drift-chicane}), motion blur (\textit{bmx-trees}, \textit{dance-twirl}, and \textit{motocross-jump}), and occlusion (\textit{horsejump-high}, \textit{kite-surf}, and \textit{libby}). The ground truth is shown in the top row, and our results are shown in the bottom row.}
	\label{fig:davis_qual}
\end{figure*}

To widely discuss the speed-accuracy trade-offs in online methods, we show the detailed results in Table \ref{tab:SoTA}, with seven online UVOS methods, \emph{e.g.}, FSEG \cite{FSEG}, SAGE \cite{SAGE}, SFM \cite{SFM}, and UOVOS \cite{UOVOS}, taken from the VOS benchmark.
\vspace{5pt}

\noindent\textbf{\DAVIS~\texttt{val}.} As shown in Table \ref{tab:SoTA}, our method achieves the best performance among all of the online algorithms in terms of mean $\mathcal{J}$. Compared to the second-best method UOVOS \cite{UOVOS} which uses the pre-trained Mask R-CNN \cite{MaskRCNN} to remove the moving background regions, our model achieves a gain of \textbf{0.8}\% in mean $\mathcal{J}$. It is worth noting that our model relies on an auto-encoder without any other additional neural network structures to implement online subspace clustering for the UVOS. In terms of runtime efficiency, SFM \cite{SFM} is the only faster online segmentation method implemented in C++. We achieve a much higher region similarity (\textbf{22.2}\%) while being faster. 

We also show qualitative results in Fig. \ref{fig:davis_qual}. We choose some videos from the \DAVIS\ dataset with the cases of dynamic background, motion blur, and occlusion. It can be seen that our model can handle different challenges. For example, our method can segment foreground objects when they are occluded by the background, as shown in the occlusion case in Fig. \ref{fig:davis_qual}. When a similar object distractor appears (\emph{e.g.}, the crowd in \textit{breakdance}, or the small camel in \textit{camel}), our method is able to discriminate a foreground target from background distractors.
\vspace{5pt}

\noindent\textbf{\FBMS~\texttt{val}.} As shown in Table \ref{tab:SoTA}, our method significantly outperforms all previous published works on the \FBMS\ \texttt{val} set compared to online UVOS methods. For instance, on the mean $\mathcal{J}$ metric, our method surpasses UOVOS \cite{UOVOS} by \textbf{2.9}\% and SAGE \cite{SAGE} by \textbf{5.2}\%. In comparison to the performance on \DAVIS, our method has a certain gap (\emph{i.e.}, \textbf{8.6}\%) on the \FBMS\ dataset. This is because our method relies only on the optical flow, and some sequences of the \FBMS\ dataset contain multiple objects in a single video. In these challenging videos, only a subset of objects are moving, so it is difficult to determine all the objects by optical flow without considering other cues.
\vspace{5pt}

\noindent\textbf{\STv~\texttt{val}.} We also report the performance of the low-resolution dataset in Table \ref{tab:SoTA}. Compared to the high-resolution \DAVIS\ dataset, it is more difficult to train an accurate optical flow model on the \STv\ dataset. Our algorithm outperforms the online methods, \emph{i.e.}, SAGE \cite{SAGE}, FSEG \cite{FSEG}, and UOVOS \cite{UOVOS}, by \textbf{5.0}\%, \textbf{1.2}\%, and \textbf{1.1}\%, respectively, in terms of mean $\mathcal{J}$, respectively. However, compared to the high-resolution datasets (\emph{i.e.}, \DAVIS\ and \FBMS), our method performs worse on the \STv\ dataset for the following reasons: 1) we have only grouped the pixels that possess the same motion pattern based on the optical flow, which significantly limits the model in segmenting objects when the flow is incomplete; and 2) some of the low-resolution videos from the \STv\ dataset affect the performance of our model, which only uses the optical flow as its input. This effect is also seen in FSEG \cite{FSEG}, SAGE \cite{SAGE}, and UOVOS \cite{UOVOS}. In particular, since UOVOS \cite{UOVOS} detects the foreground object based on a salient motion map, using Mask R-CNN on incomplete optical flow provides little improvement.

\subsection{Runtime Comparison}\label{Exp:Run}

To further investigate the computational efficiency of our proposed method, we report the inference time comparisons on the \DAVIS\ datasets at 480p resolution. We compare our model with the SoTA online methods that share their codes or include the corresponding experimental results, including the SFM \cite{SFM}, and UOVOS \cite{UOVOS}. For the inference time comparison, we run the public code of other methods and our code under the same conditions on the NVIDIA TITAN RTX GPU. The analysis results are summarized in the last column of Table \ref{tab:SoTA}.

As shown in Table \ref{tab:SoTA}, our algorithm shows a faster speed than other competitors. For online UVOS settings, model efficiency is an important metric. Our model achieves a more favorable accuracy-efficiency trade-off than the existing best online method UOVOS \cite{UOVOS}, while achieving higher accuracy. The main computational cost of UOVOS \cite{UOVOS} lies in the object proposal component, which is based on Mask R-CNN \cite{MaskRCNN}. However, our model relies on an auto-encoder and online clustering strategy for the UVOS without any other additional neural network structures. Compared to the faster online method SFM \cite{SFM}, our model achieves a \textbf{22.2}\% higher mean $\mathcal{J}$.

\section{Conclusion Remarks and Discussions}
\label{sec:con}
In this paper, an efficient contrastive subspace motion clustering is proposed for online unsupervised video object segmentation (UVOS) by exploring an online clustering strategy for motion grouping. Specifically, non-learnable prototypical bases are iteratively summarized from the feature space for different motion patterns, and these bases help to optimize the feature representation in return. Experimental results demonstrated that our method outperforms state-of-the-art (SoTA) online UVOS algorithms.

In real-world scenarios, the performance of our online UVOS system may be violated due to the presence of low-resolution input, making it inaccurate for small objects. Therefore, we can see that our method performs worse on the low-resolution dataset, \emph{i.e.}, \STv, compared to the high quality video data. The saturation of moving objects, such as white vehicles under sunshine and black vehicles in dim lighting conditions, will also affect the proposed UVOS method. The problem will be investigated by utilizing the results in \cite{DSRGAN} and \cite{Yilun, Chaobing}, respectively. Furthermore, considering the significance of real-time aspects in online UVOS, we will incorporate a light-weighted design \cite{DSN, RTLWUVOS} in our future research.

\bibliographystyle{IEEEtran}
\bibliography{mybib}

\vfill

\end{document}